\documentclass[accepted]{uai2025} 
                        
\usepackage[american]{babel}

\usepackage{natbib} 
\usepackage{mathtools} 
\usepackage{booktabs} 
\usepackage{tikz} 

\title{Efficiently Escaping Saddle Points for Policy Optimization}

\author[1]{\href{mailto:<sadegh.khorasani@epfl.ch>?Subject=Your UAI 2025 paper}{Sadegh Khorasani}{}} 
\author[2]{\href{mailto:<s.salehkaleybar@liacs.leidenuniv.nl>?Subject=Your UAI 2025 paper}{Saber Salehkaleybar}{}}
\author[3]{\href{mailto:<negar.kiyavash@epfl.ch>?Subject=Your UAI 2025 paper}{Negar Kiyavash}{}}
\author[4]{\href{mailto:<niao.he@inf.ethz.ch>?Subject=Your UAI 2025 paper}{Niao He}{}}
\author[1]{\href{mailto:<matthias.grossglauser@epfl.ch>?Subject=Your UAI 2025 paper}{Matthias Grossglauser}{}}

\affil[1]{%
    School of Computer and Communication Sciences, EPFL,\\ Lausanne, Switzerland
}

\affil[2]{%
    Leiden Institute of Advanced Computer Science (LIACS), Leiden University, \\Leiden, The Netherlands
  }

\affil[3]{%
    College of Management of Technology, EPFL, Lausanne, Switzerland
}

\affil[4]{%
  Department of Computer Science ETH Zurich
}

\usepackage{enumitem}
\usepackage{hyperref}
\usepackage{url}
\usepackage{graphicx}
\usepackage{subfigure}
\usepackage{paralist}
\usepackage{amsmath}
\usepackage{amssymb}
\usepackage{amsthm}

\usepackage[capitalize,noabbrev]{cleveref}
\usepackage{verbatim}
\usepackage{subfigure}
\usepackage{booktabs}
\usepackage{marvosym}
\usepackage{amssymb}
\usepackage{cases}
\usepackage{mwe}
\usepackage{paralist}
\usepackage{mathtools}
\usepackage{algorithm}
\usepackage{algorithmic}
\usepackage{enumerate}      
\usepackage{wrapfig}
\usepackage{adjustbox}
\let\classAND\AND
\let\AND\relax
\usepackage{algorithmic}

\let\AND\classAND
\AtBeginEnvironment{algorithmic}{\let\AND\algoAND}

\usepackage{hyperref}

\theoremstyle{plain}
\newtheorem{theorem}{Theorem}[section]

\newtheorem{lemma}[theorem]{Lemma}
\newtheorem{corollary}[theorem]{Corollary}
\theoremstyle{definition}
\newtheorem{definition}[theorem]{Definition}
\newtheorem{assumption}[theorem]{Assumption}
\theoremstyle{remark}
\newtheorem{remark}[theorem]{Remark}

\newcommand{\OurFloor}[1]{\lfloor #1 \rfloor}
  
\begin{document}
\maketitle

\begin{abstract}
Policy gradient (PG) is widely used in reinforcement learning due to its scalability and good performance. In recent years, several variance-reduced PG methods have been proposed with a theoretical guarantee of converging to an approximate first-order stationary point (FOSP) with the sample complexity of $O(\epsilon^{-3})$. However, FOSPs could be bad local optima or saddle points. Moreover, these algorithms often use importance sampling (IS) weights which could impair the statistical effectiveness of variance reduction. In this paper, we propose a variance-reduced second-order method that uses second-order information in the form of Hessian vector products (HVP) and converges to an approximate second-order stationary point (SOSP) with sample complexity of $\tilde{O}(\epsilon^{-3})$. This rate improves the best-known sample complexity for achieving approximate SOSPs by a factor of $O(\epsilon^{-0.5})$. Moreover, the proposed variance reduction technique bypasses  IS weights by using HVP terms. Our experimental results show that the proposed algorithm outperforms the state of the art and is more robust to changes in random seeds.
\end{abstract}

\section{Introduction}
\label{sec:intro}
Reinforcement Learning (RL) is an interactive learning approach where an agent learns how to choose the best action in an interactive environment based on received signals. In the past few years, RL has been applied with great success to many applications of interest such as autonomous driving \citep{shalev2016safe}, games \citep{silver2017mastering}, and robot manipulation \citep{deisenroth2013survey}. RL can be formulated mathematically as a Markov Decision Process (MDP) where after taking an action, the state changes based on transition probabilities and the agent receives a reward according to the current state and the action taken. The agent takes actions according to a policy that maps the state space to action space, and the goal is to find a policy that maximizes the agent's average cumulative reward. 

Policy gradient (PG) methods directly search for a policy that is parameterized by a parameter vector $\theta$. PG methods can provide good policies in high-dimensional control tasks by first harnessing the power of deep neural networks for policy parameterization and subsequently optimizing the parameter vector $\theta$. REINFORCE \citep{williams1992simple}, PGT \citep{sutton2000policy}, and GPOMDP \citep{baxter2001infinite} are examples of classical PG methods that update the parameters using stochastic gradient ascent, as it is often infeasible to compute the gradient exactly in complex environments. 

{ To manage the stochasticity in gradient update, in the RL literature, several approaches have been proposed.  \citet{sutton2000policy} presented the baseline technique to reduce the variance of gradient estimates.} \citet{konda2000actor} proposed an actor-critic algorithm that estimates the value function and utilizes it for the purpose of variance reduction. \citet{schulman2015high} presented GAE to control both bias and variance by exploiting a temporal difference relation for the advantage function approximation. \citet{schulman2015trust} proposed TRPO, which considers a Kullback-Leibler (KL) divergence penalty term in order to ensure that the updated policy remains close to the current policy. Subsequently, \citet {schulman2017proximal}
used a clipped surrogate objective function. 
It has been observed experimentally that the aforementioned algorithms have better performance compared to the vanilla PG method. However, no theoretical guarantees for the convergence rate of most of these algorithms are available.

In recent years, several methods have been proposed with theoretical guarantees for the convergence to an $\epsilon$-approximate First-Order Stationary Point ($\epsilon$-FOSP) of the objective function $J(\theta)$, i.e., $\| \nabla J(\theta)\| \leq \epsilon$. Most of these methods adopt recent variance reduction techniques presented originally in the context of stochastic optimization, and converge to $\epsilon$-FOSP with the sample complexity of $O(\epsilon^{-3})$ (See Related Work section for more details). The two main drawbacks of these methods are as follows. First, most of these methods require importance sampling (IS) weights in the variance reduction part because the objective function in RL is non-oblivious in the sense that its trajectories depend on the policy that generates them. This degrades the effectiveness of variance reduction techniques, because the IS weights grow exponentially with the horizon length \citep{zhang2021convergence}. Moreover, their analyses required strong assumptions such as the boundedness of variance of IS weights. The second drawback is the failure to provide guarantees beyond convergence to $\epsilon$-FOSP. In many applications, the objective function $J(\theta)$ is non-convex and FOSPs may include bad local optima and saddle points. This is one of the reasons why the aforementioned methods are too sensitive to parameter initialization and random seeds in practice.
{ We argue that it is more desirable to obtain  $(\epsilon,\sqrt{\rho\epsilon})$-approximate second-order stationary point ($(\epsilon,\sqrt{\rho\epsilon})$-SOSP), i.e., $\| \nabla J(\theta)\| \leq \epsilon$ and $\lambda_{max}(\nabla^2 J(\theta))\leq \sqrt{\rho\epsilon}$, where $\lambda_{max}(.)$ is the maximum eigenvalue and $\rho$ is the Hessian Lipschitz constant\footnote{The function \( J \) is said to have a Hessian Lipschitz constant \( \rho \) if for all \( \theta_1, \theta_2  \in \mathbb{R}^{d} \): $\| \nabla^2 J(\theta_1) - \nabla^2 J(\theta_2) \| \leq \rho \| \theta_1 - \theta_2 \|
$.}. }

In the context of optimization, \cite{nesterov2006cubic} showed that cubic regularized Newton (CRN) method escapes saddle points and converges to SOSP by incorporating second-order information, namely, the Hessian matrix. However, obtaining the  Hessian is computationally intensive in high dimensions, which is the case in most RL applications where the policy is modeled by a (deep) neural network. Moreover, in the stochastic setting, only estimates of the gradient and Hessian are available. To address these challenges, \citet{tripuraneni2018stochastic} proposed stochastic CRN (SCRN), which uses sub-sampled gradient and Hessian to converge to $(\epsilon,\sqrt{\rho\epsilon})$-SOSP with sample complexity of $\tilde{O}(\epsilon^{-3.5})$\footnote{$\tilde{O}(.)$ is a variant of big-$O$ notation, ignoring the logarithmic factors. In other words, $f(n)\in \tilde{O}(g(n))$ if there exists some positive constant $k$, such that $f(n)\in O(g(n)\log^k(g(n)))$ .}. Furthermore, they showed that using the result in \citep{carmon2016gradient}, the update in each iteration can be obtained from Hessian vector products (HVP) instead of computing the full Hessian matrix. 
 Later, \citet{zhou2020stochastic} proposed a variance-reduced version of SCRN based on SARAH \citep{nguyen2017sarah} achieving sample complexity of $\tilde{O}(\epsilon^{-3})$.

{In the context of RL, recently, \citet{wang2022stochastic} proposed a second-order PG method that converges to $(\epsilon,\sqrt{\rho\epsilon})$-SOSP with a sample complexity of $\tilde{O}(\epsilon^{-3.5})$. The proposed method uses a variance reduction technique based on SARAH to estimate the gradients. This technique still requires IS weights and the customary strong assumptions on these weights (such as the boundedness of their variance). Concurrent to \citet{wang2022stochastic}, \citet{maniyar2024cubic}, proposed a second-order PG method that converges to $(\epsilon,\sqrt{\rho\epsilon})$-SOSP with the best-known sample complexity rate of ${O}(\epsilon^{-3.5})$.
A natural question to ask is whether there exists a second-order PG method that converges to$(\epsilon,\sqrt{\rho\epsilon})$-SOSP with sample complexity of $\Tilde{O}(\epsilon^{-3})$ but without using IS weights?}

In this paper, we answer the above question in the affirmative by proposing Variance-Reduced Stochastic Cubic-regularized Policy gradient (VR-SCP) algorithm.  The proposed algorithm updates the parameters in each iteration based on optimizing a stochastic second-order Taylor expansion of the objective function with a cubic penalty term where gradient estimates are obtained based on a novel variance reduction technique. In particular, our main contributions are as follows:
\begin{compactitem}
    \item VR-SCP converges to $(\epsilon,\sqrt{\rho\epsilon})$-SOSP with sample complexity of $\tilde{O}(\epsilon^{-3})$, improving the best-known sample complexity \citep{wang2022stochastic, maniyar2024cubic} by a factor of $O(\epsilon^{-0.5})$.
    \item We propose a Hessian-aided variance reduction technique that incorporates HVP in estimating gradients, entirely bypassing IS weights. Our convergence analysis does not require strong assumptions on IS weights.

    \item To showcase the advantages of converging to $(\epsilon,\sqrt{\rho\epsilon})$-SOSP, we define a new metric that incorporates both 
    performance (the average return) and robustness (sensitivity to random seeds) of an RL algorithm, where the latter is crucial in terms of reproducibility of the results. Our experimental results show that not only VR-SCP outperforms state-of-the-art algorithms in terms of its theoretical guarantees, but also in terms of the aforementioned metric.

\end{compactitem}



 We should emphasize that our technique differs from  HAPG \citep{shen2019hessian} which also bypasses IS weights using HVP in 
 three main aspects. First, in their analysis, it is required to update the parameters with a fixed step size of $\epsilon$ in order to bound the variance of gradient estimates which slows the training process in practice. Second, in HAPG, the number of computed HVPs per iteration is in the order of $O(1/\epsilon)$ while in our case, it depends on the norm of the update. Last but not least, HAPG only uses the second-order information in the form of HVP for variance reduction and hence achieves $\epsilon$-FOSP, therefore, as we shall see in our experiments, it misses the main performance advantages of converging to $(\epsilon,\sqrt{\rho\epsilon})$-SOSP. 
 
The paper is organized as follows: In Section 2, we provide some definitions and background on variance-reduced methods in stochastic optimization and also the second-order method, SCRN. In Section 3, we describe the proposed algorithm and analyze its convergence rate for achieving $(\epsilon,\sqrt{\rho\epsilon})$-SOSP. In Section 4, we review related work in the RL literature. In Section 5, we define a new metric to evaluate RL algorithms in control tasks and compare the proposed algorithm with previous work based on this metric. Finally, we conclude the paper in Section 6.

\section{Preliminaries}
\label{sec:pre}
\subsection{Notations and problem definition}

An MDP can be represented as a tuple $\mathcal{M}=\langle\mathcal{S},\mathcal{A},P,R, \gamma,p_0\rangle$, where $\mathcal{S}$ and $\mathcal{A}$ are state space and action space, respectively. 
The conditional probability of transition from state $s$ to $s'$ with action $a$ is denoted by $P(s'|s,a)$. 
The probability distribution over the initial state $s_0$ is denoted by $p_0(s_0)$. 
The parameter $\gamma\in (0,1)$ denotes the discount factor.
At each time step $t$, $r(s_t,a_t)$ returns the reward of taking action $a_t$ in the state $s_t$. 
Actions are chosen according to the policy $\pi$ where $\pi(a|s)$ is the probability of taking action $a$ for a given state $s$. Here, we assume that the policy is parameterized with a vector $\theta\in \mathbb{R}^d$ and use shorthand notation $\pi_{\theta}$ for $\pi_{\theta}(a|s)$. For a given time horizon $H$, we define $\tau=(s_0,a_0,\cdots,s_{H-1},a_{H-1})$ as a sequence of state-action pairs called a trajectory. $R(\tau)$ is a function that returns the discounted accumulated reward of each trajectory as follows:
$R(\tau):=\sum_{h=0}^{H-1} \gamma^h r(s_h,a_h)$ where $\gamma\in (0,1)$ is the discount factor. For an arbitrary vector $v \in \mathbb{R}^d$, we denote Euclidean norm by $||v||_2$. 
For a matrix $A \in \mathbb{R}^{a \times b} $, $A[.]: \mathbb{R}^b \rightarrow \mathbb{R}^a $ takes a vector $v\in \mathbb{R}^b$ and returns the matrix-vector product $Av$. 
We show the maximum eigenvalue of a symmetric matrix $H \in \mathbb{R}^{d \times d}$ with $\lambda_{max}(H)$. 
For two vectors $v_1, v_2  \in \mathbb{R}^d$, we use $\langle v_1, v_2\rangle$ to denote their vector product. The spectral norm (\(\|A\|_2\)) of a matrix \(A\) is defined as the square root of the largest eigenvalue of the symmetric matrix \(A^TA\), where \(A^T\) is the transpose of matrix \(A\). Throughout the rest of the paper, we may omit the subscript in the norm notation for the sake of brevity.

Variance-reduced methods for estimating the gradient vector were originally proposed for the stochastic optimization setting, i.e.,
\begin{equation}
    \min_{\theta\in \mathbb{R}^d} F(\theta)=\mathbb{E}_{z\sim p(z)} [f(\theta,z)],
\end{equation}
where a sample $z$ is drawn from the distribution $p(z)$ and $f(.,z)$s are commonly assumed to be smooth and non-convex functions of $\theta$. This setting is mainly considered in the supervised learning context where $\theta$ corresponds to the parameters of the training model and $z=(x,y)$ is the training sample, where $x$ denotes the feature vector of the sample and $y$ is the corresponding label. In this setting, the distribution $p(z)$ is invariant with respect to parameter $\theta$.

In the RL setting, the goal is to maximize the expected cumulative reward:
\begin{equation}
    \label{eq:obj_rl}
    \max_{\theta\in \mathbb{R}^d} J(\theta)=\mathbb{E}_{\tau\sim \pi_{\theta}}[R(\tau)],
\end{equation}

where $\theta$ corresponds to the parameters of the policy and trajectory $\tau$ is drawn from distribution $\pi_\theta$. 
The probability of observing a trajectory $\tau$ for a given policy $\pi_{\theta}$ is:
\begin{equation}
    p(\tau|\pi_{\theta})=p_0(s_0)\prod_{h=0}^{H-1}P(s_{h+1}|s_h,a_h)\pi_{\theta}(a_h|s_h).
\end{equation}

Unlike supervised learning, the distribution of these trajectories depends on the parameters of policy $\pi_\theta$. 
It can be shown that:
\begin{equation}
    \nabla J(\theta)=\mathbb{E}_{\tau\sim \pi_{\theta}}\Bigg[\sum_{h=0}^{H-1} \Psi _h(\tau) \nabla \log\pi_{\theta}(a_h|s_h)\Bigg],
\end{equation}
where $\Psi_h(\tau)=\sum_{t=h}^{H-1} \gamma^t r(s_t,a_t)$. Therefore, for any trajectory $\tau$, $\hat{\nabla} J(\theta,\tau):=\sum_{h=0}^{H-1} \Psi _h(\tau) \nabla \log\pi_{\theta}(a_h|s_h)$ is an unbiased estimator of $\nabla J(\theta)$. The vanilla policy gradient updates $\theta$ as follows:
\begin{equation}
    \theta\gets \theta+\eta \hat{\nabla} J(\theta, \tau),
\end{equation}
where $\eta$ is the learning rate. 

The Hessian matrix of $J(\theta)$ can be written as follows \citep{shen2019hessian}:
{
\begin{equation}
    \nabla^2 J(\theta)=\mathbb{E}_{\tau\sim \pi_{\theta}}[\nabla \Phi(\theta,\tau)\nabla \log p(\tau|\pi_{\theta})^T+ \nabla^2\Phi(\theta,\tau)],
    \label{eq:Hessian}
\end{equation}}
where $\Phi(\theta,\tau)=\sum_{h=0}^{H-1}\sum_{t=h}^{H-1} \gamma^t r(s_t,a_t) \log \pi_{\theta}(a_h|s_h)$ for a given trajectory $\tau$.
Hence, $\hat\nabla^2J(\theta,\tau):=\nabla \Phi(\theta,\tau)\nabla \log p(\tau|\pi_{\theta})^T+ \nabla^2\Phi(\theta,\tau)$ is an unbiased estimator of the Hessian matrix.
Note that the computational complexity of the term $\nabla \Phi(\theta,\tau)\nabla \log p(\tau|\pi_{\theta})^T$ is $O(Hd)$ where $d$ is the dimension of the gradient vector. The second term is HVP which can also be computed in $O(Hd)$ using Pearlmutter's algorithm \citep{pearlmutter1994fast}.

We can compute HVP using the above definition for the sample-based version of Hessian matrix $\hat{\nabla}^2J(\theta,\tau)$ and an arbitrary vector $v \in \mathbb{R}^d$ as follows:
\begin{equation}
    \hat{\nabla}^2J(\theta,\tau)[v]=(\nabla \log p(\tau|\pi_{\theta})^Tv)\nabla \Phi(\theta,\tau)+ \nabla^2\Phi(\theta,\tau)[v].
\end{equation}

\subsection{Variance reduced methods for gradient estimation}
For any time $ t \geq 1 $ and a sequence of parameters 
$\{\theta_0,\theta_1,\cdots\}$, we can write the gradient at $\theta_t$ as follows: 
\begin{equation}
\nabla J(\theta_{t}) = \nabla J(\theta_{t-1}) + \nabla J(\theta_{t})- \nabla J(\theta_{t-1}).
\label{eq:case0}
\end{equation}
Suppose that we have an unbiased estimate of $\nabla J(\theta_{t-1})$ at time $t-1$, denoted by $v_{t-1}$. If we have an unbiased estimate of $\nabla J(\theta_{t})-\nabla J(\theta_{t-1})$ denoted by $\Delta_t$, then we can add it to $v_{t-1}$ in order to get an unbiased estimate of $\nabla J(\theta_{t})$ at time $t$ as follows:
\begin{equation}
v_{t} = v_{t-1} + \Delta_t.
\label{eq:case1}
\end{equation}

Let $\epsilon_t = v_t -\nabla J(\theta_t)$ and $\epsilon_{\Delta_t}=\Delta_t - (\nabla J(\theta_{t})- \nabla J(\theta_{t-1}))$. Based on these definitions, we can rewrite the above equation as follows:
\begin{equation}
    \epsilon_t = \epsilon_{t-1}+\epsilon_{\Delta_t}.
\end{equation}
Thus, we have:
\begin{equation}
    \mathbb{E} [\|\epsilon_t\|^2] = \mathbb{E} [\|\epsilon_{t-1}\|^2] + \mathbb{E} [\|\epsilon_{\Delta_t}\|^2] + 2 \mathbb{E}[\langle\epsilon_{t-1},\epsilon_{\Delta_t}\rangle].
\end{equation}
The above equation shows that if $\mathbb{E}[\langle\epsilon_{t-1},\epsilon_{\Delta_t}\rangle]$ is sufficiently negative, then $\mathbb{E} [\|\epsilon_t\|^2]$
is decreasing in time.
The updates in several previous variance reduction methods (such as in HAPG \citep{shen2019hessian}, MBPG \citep{huang2020momentum}, SRVR-PG \citep{xu2019sample}, SVRG \citep{johnson2013accelerating} and SARAH \citep{nguyen2017sarah}) are consistent with \eqref{eq:case1} with different suggestions for $\Delta_t$.
{
Using the update in \eqref{eq:case1} may accumulate errors in the gradient estimates and some of the methods in the literature require checkpoints after some iterations and use a batch of stochastic gradients at these points to control the error.
}



\subsection{Stochastic Cubic Regularized Newton}


In the context of optimization, Cubic Regularized Newton (CRN) \citep{nesterov2006cubic} obtains SOSP in general non-convex functions  by updating the parameters using the cubic-regularized second-order expansion of the Taylor series at each iteration $t$ as follows:
\begin{align}
    &m_t(h) = \langle {\nabla F(\theta_t)}, {h}\rangle+\frac{1}{2}\langle  {\nabla^2 F(\theta_t)h}, {h} \rangle+\frac{M}{6}\|{h}\|^{3},\\
    &h^*_t = {\bf argmin}_{{h}\in\mathbb{R}^{d}} m_t(h),\\
    & \theta_{t+1} = \theta_t + h^*_t.\nonumber
\end{align}

In CRN, the sub-problem for obtaining the minimizer of $m_t(h)$, should be solved in each iteration. Although there is no closed-form solution for the sub-problem, \cite{carmon2016gradient} proposed a gradient descent-based algorithm to find an approximate solution.

As we often only have access to gradient and Hessian estimates of the objective function (herein, denoted by $v_t$ and $U_t$, respectively), the aforementioned update rule in the context of RL becomes:

\begin{align}
    &m_t(h) = \langle {v_t}, {h}\rangle+\frac{1}{2}\langle {U_th}, {h} \rangle-\frac{M}{6}\|{h}\|^{3},\label{eq:body_sub_problem}\\
    &h^*_t ={\bf argmax}_{{h}\in\mathbb{R}^{d}} m_t(h),\label{eq:body_arge_max_sub_problem}\\
    & \theta_{t+1} = \theta_t + h^*_t.
\end{align}

\section{VR-SCP Algorithm}
\label{sec:algo}

In this section, we first describe our proposed algorithm, VR-SCP, which uses the second-order information in order to have a better estimate of the gradient vector and converges to $(\epsilon,\sqrt{\rho\epsilon})$-SOSP. Next, we provide a convergence analysis of the proposed algorithm under some customary regularity assumptions in RL literature.

\subsection{Description}

The algorithm iterates in a loop starting from line 1. 
In line 2, the estimate of gradient, $v_t$, is computed. After every $Q$ iterations, we have a checkpoint (when the condition $\mod(t,Q)=0$ is satisfied) where $v_t$ is set to the average of a batch of stochastic gradients.
Otherwise, based on what we explained in \eqref{eq:case1}, the following term will be added to the last gradient estimate $v_{t-1}$, as an estimate of $\nabla J(\theta_t)-\nabla J(\theta_{t-1})$:
\begin{equation}
\dfrac{1}{S_t}\sum_{s=1}^{S_t} \hat{\nabla}^2 J(\theta_{s,t},\tau_s)(\theta_t-\theta_{t-1}),  
\end{equation}
 where parameter $\theta_{s,t}$ is a point on the line between $\theta_{t-1}$ and $\theta_t$ and it is equal to $\left(1-\dfrac{s}{S_t}\right)\theta_{t}+\dfrac{s}{S_t}\theta_{t-1}$. Furthermore, $S_t$ is the number of points taken on the line, and $\tau_s$ is a trajectory drawn based on policy $\pi_{\theta_{s,t}}$.  In line 3, VR-SCP defines a stochastic HVP function $U_{t}[.]$ which for every vector $v$, it computes $U_{t}[v] \gets \frac{1}{|\mathcal{B}_{h}|}\sum_{\tau\in \mathcal{B}_{h}}\hat{\nabla}^{2} J(\theta_{t},\tau)[v] $.
In line 4, Cubic-Subsolver runs a gradient descent-based algorithm to find an approximate solution of the sub-problem. In line 5, if there is a sufficient increase in $m_t(h)$, (i.e., $m_t({h}_{t})\geq  {\rho}^{-1/2}\epsilon^{3/2}/6$), the parameter $\theta_t$ is updated based on the approximate solution $h_t$. Otherwise, Cubic-Finalsolver is called once and the algorithm terminates after updating the parameter $\theta_t$.
Cubic-Subsolver and Cubic-Finalsolver are based on \citep{tripuraneni2018stochastic} and the pseudo-codes of these two algorithms are given in Appendix \ref{app:subsolver}.

\begin{algorithm}[t]
\caption{Variance-reduced stochastic cubic regularized Newton-based policy gradient (VR-SCP)}\label{alg:cap}
\textbf{Input:} Batch $\mathcal{B}_{check}$ and $\mathcal{B}_{h}$,  initial point $\theta_{0}$, accuracy $\epsilon$, cubic penalty parameter $M$,  maximum number of iterations $T$, duration $Q$, and parameters $L, \rho$.
\begin{algorithmic}[1]
\FOR{$t=0,\cdots,T-1$}
    \STATE
    \small
    \hspace{-3cm}
    \begin{numcases}{v_t=}
        \frac{1}{|\mathcal{B}_{check}|}\sum_{\tau\in \mathcal{B}_{check}} \hat{\nabla} J(\theta_t,\tau), & \hspace{-1.75cm} $\text{if mod}(t,Q)=0$,  \nonumber\\
        \dfrac{1}{S_t}\sum_{s=1}^{S_t} \hat{\nabla}^2 J(\theta_{s,t},\tau_s)(\theta_t-\theta_{t-1})+ v_{t-1} ,  & o.w. \nonumber
    \end{numcases}
    \STATE $U_{t}[.] \gets \frac{1}{|\mathcal{B}_{h}|}\sum_{\tau\in \mathcal{B}_{h}}\hat{\nabla}^{2} J(\theta_{t},\tau)[.] $
    \STATE ${h}_{t} \gets \textbf{Cubic-Subsolver}(U_t[.] , v_t, M, L, \epsilon)$
    \IF {$m_t({h}_{t}) > {\rho}^{-1/2}\epsilon^{3/2}/6$}
            \STATE $\theta_{t+1}\gets \theta_{t}+{h}_{t}$
        \ELSE 
            \STATE ${h}_{t} \gets \textbf{Cubic-Finalsolver}(U_t[.] , v_t, M, L, \epsilon)$
            \STATE \textbf{return} $\theta_{t+1}\gets \theta_{t}+{h}_{t}$
        \ENDIF
    \STATE $t \gets t+1$
\ENDFOR
\STATE \textbf{return} $\theta_{T}$
\end{algorithmic}
\end{algorithm}

\subsection{Convergence Analysis}
In this part, we analyze the convergence rate of the proposed algorithm under the bounded reward function and some common regularity assumptions on the policy.

\begin{assumption}
[Bounded reward] For $\forall s\in \mathcal{S}, \forall a \in \mathcal{A}$, $|R(s,a)|<R_0$ where $R_0>0$ is some constant.
\label{assum:1}
\end{assumption}
{
\begin{assumption}
[Parameterization regularity]
There exist constants $G,L_1,L_2>0$ such that for any $\theta_1,\theta_2,w\in \mathbb{R}^d$ and for any $ s\in \mathcal{S}, a \in \mathcal{A}$:
\\
(a) $\|\nabla \log\pi_{\theta_1}(a|s)\|\leq G$, \\
(b) $\|\nabla^2 \log\pi_{\theta_1}(a|s)\|\leq L_1$, \\
(c) $\|(\nabla^2 \log\pi_{\theta_1}(a|s)-\nabla^2 \log\pi_{\theta_2}(a|s))w\|\leq L_2 \|\theta_1-\theta_2\|\|w\|$.

\label{assum:2}
\end{assumption}
}

Assumptions \ref{assum:1} and \ref{assum:2} (a,b), commonly used in the RL literature to analyze the convergence of policy gradient methods, imply boundedness of the norm of the stochastic gradient as well as the individual smoothness of the stochastic gradient and Hessian. 
{
\begin{lemma}
\label{lem:lipschizness}
\citep{shen2019hessian,wang2022stochastic} Under Assumptions \ref{assum:1} and \ref{assum:2}, for any $\theta_1,\theta_2, w\in \mathbb{R}^d$ and for any trajectory $\tau$, there exist constants $W$, $L$ and $\rho$ such that:
\begin{equation*}
\begin{split}
    &\|\hat{\nabla} J(\theta_1,\tau)\|_{2}\le W,\\
    &\|\hat{\nabla} J(\theta_1, \tau)-\hat{\nabla} J(\theta_2,\tau)\|_{2}\le L\|\theta_1-\theta_2\|_{2},\\
    &\|\hat{\nabla}^{2} J(\theta_1,\tau)-\hat{\nabla}^{2} J(\theta_2,\tau)\|\le \rho\|\theta_1-\theta_2\|_{2},
\end{split}
\end{equation*}
where $\rho=\frac{L_2 R_0+2R_0GHL_1}{(1-\gamma)^2}, L= \frac{L_1GR_0}{(1-\gamma)^2} , W = \frac{GR_0}{(1-\gamma)^2}$.
\end{lemma}
}

\begin{remark}
\label{rem:hessian_lipchitz}
Assumption \ref{assum:2} (c) is also a common in analyzing second-order methods in RL. Examples of policies that satisfy this assumption are Gaussian \citep{pirotta2013adaptive} and soft-max policies \citep{masiha2022stochastic}. For instance, consider a Gaussian policy 
with standard deviation $\sigma$ as follows:
\begin{equation*}
\pi_{\theta}(a|s)=\mathcal{N}(\theta^T\mu(s),\sigma^2),
\end{equation*}
where $\mu(s):\mathcal{S}\rightarrow \mathbb{R}^d$ is a feature map. It can be easily seen that $\nabla^2\log\pi_{\theta}(a|s)=\mu(s)^T\mu(s)/\sigma^2$. Therefore, Gaussian policy satisfies Assumption \ref{assum:2} if $\mu(s)$, $\theta$, and actions take values in a bounded domain.
\end{remark}

Assumptions \ref{assum:1} and \ref{assum:2} also imply that the variance of gradient estimate $\hat{\nabla} J(\theta,\tau)$ and Hessian estimate $\hat{\nabla}^2 J(\theta,\tau)$ are bounded.

\begin{lemma} \citep{shen2019hessian} Under Assumptions \ref{assum:1} and \ref{assum:2}, for any point $\theta \in \mathbb{R}^d$ and trajectory $\tau$, $\hat{\nabla} J(\theta,\tau)$ and $\hat{\nabla}^2 J(\theta,\tau)$, have bounded variances $\sigma_1^2$ and $\sigma_2^2$, respectively:
\begin{equation}
\begin{split}
    &\mathbb{E}[\|\hat{\nabla} J(\theta,\tau)-\nabla J(\theta)\|]^2\le \sigma^{2}_{1}\\
    &\mathbb{E}[\|\hat{\nabla}^2 J(\theta,\tau)-\nabla^2 J(\theta)\|^{2}]\le \sigma^{2}_{2},
\end{split}
\end{equation}

where $\sigma_1^2=\frac{G^2R_0^2}{(1-\gamma)^4}$ and $\sigma_2^2=\frac{H^2G^4R_0^2+L_1^2R_0^2}{(1-\gamma)^4}$.

\end{lemma}

Following the work \citep{nesterov2006cubic} on cubic regularized Newton method, we define the following quantity for showing convergence to $(\epsilon,\sqrt{\rho\epsilon})$-SOSP. 
\begin{definition}
\label{def:inj}
For any $\theta\in \mathbb{R}^d$, we define $\mu(\theta)$ as follows:
\begin{equation}
    \mu(\theta)= \max(\|\nabla J(\theta)\|^{3/2}, \lambda^3_{max}(\nabla^{2}J(\theta))/\rho^{3/2}).
\end{equation}
\end{definition}
Based on the above definition, it can be easily shown that the point $\theta$ is $(\epsilon, \sqrt{\rho\epsilon})$-SOSP if and only if $\mu(\theta) \leq \epsilon^{3/2}$.

To show the convergence of the proposed algorithm, we use the following lemma which provides a bound on the norm of estimation error at each iteration $t$ with high probability given history up to time $\OurFloor{t/Q}.Q$.
{
\begin{lemma}
\label{lem:grad_bound} Let $\mathcal{F}_t$ be the history up to time $t$. Under Assumptions \ref{assum:1}, \ref{assum:2} and for the values of $S_t$ in Appendix \ref{sec:lemma err est for grad} and $|\mathcal{B}_{check}|$ in the statement of Theorem \ref{th:main}, conditioned on $\mathcal{F}_{\lfloor t/Q\rfloor.Q}$, with probability $1-2\delta(t-\lfloor t/Q \rfloor.Q )$, for all $i$ between $\lfloor t/Q\rfloor.Q$ and $t$, we have:
    \begin{align}
        \| v_i-\nabla J(\theta_i) \|^2_{2}\le \frac{\epsilon^2}{30}.
    \end{align}
\end{lemma}
All the proofs of lemmas and theorem appear in the Appendix.
}
\begin{lemma}
\label{lem:hess_bound}
For the value of $|\mathcal{B}_h|$ in the statement of Theorem \ref{th:main},
under Assumptions \ref{assum:1}, \ref{assum:2}, conditioned on $\mathcal{F}_{t}$, with probability $1-\delta$, we have:
    \begin{align}
        \| U_t-\nabla^2 J(\theta_t) \|^2\le \frac{\epsilon \rho}{30}.
    \end{align}
\end{lemma}

\begin{theorem}
\label{th:main}
Under Assumptions \ref{assum:1}, \ref{assum:2}, for  $|\mathcal{B}_{check}| \geq \dfrac{19440W^2\log^2( 4T/\xi)}{\epsilon^2} $, $|\mathcal{B}_{h}| \geq \dfrac{1080L^2\log(4dT/\xi)}{\rho\epsilon} $, and $S_t$ defined in Appendix \ref{sec:lemma err est for grad}, cubic penalty parameter $M = 4\rho$, $\epsilon \leq 4 L^2\rho/M$ and maximum number of iterations $T \geq 25\Delta_J \rho^{1/2}\epsilon^{-3/2}$, Algorithm \ref{alg:cap} guarantees that:
\begin{equation}
\begin{split}
    \mu(\theta_{out}) \leq 1300\epsilon^{3/2},
\end{split}
    \label{eq:th}
\end{equation}
with the probability $1-\xi$
where $\Delta_J =J^*-J(\theta_0)$ and $J^*$ is the optimal value of objective function.

\end{theorem}
{

\paragraph{Proof sketch.}
To prove Theorem \ref{th:main}, we establish bounds on $\| v_t - \nabla J(\theta_t) \|^2$ and $\| U_t - \nabla^2 J(\theta_t) \|^2$. A key step involves bounding the term 
\[\frac{1}{S_t}
\left\|\sum_{s=1}^{S_t} \hat{\nabla}^2 J(\theta_{s,t})(\theta_t - \theta_{t-1}) - \nabla J(\theta_t) + \nabla J(\theta_{t-1})\right\|,
\]
which we show is a quadratic function of $\|\theta_t - \theta_{t-1}\|$ with high probability (see \eqref{ub_bound} in the appendix). Using this result, we adjust the batch sizes to derive high-probability bounds on $\| v_t - \nabla J(\theta_t) \|^2$ (Lemma \ref{lem:grad_bound}) and $\| U_t - \nabla^2 J(\theta_t) \|^2$ (Lemma \ref{lem:hess_bound}). Substituting these bounds into the recursive inequality in \eqref{eq:lower_bound_of_J}, we show that the proposed algorithm achieves the convergence rate stated in Theorem \ref{th:main}.  

}

\begin{corollary}
\label{cor:main}
Under the Assumptions in the statement of Theorem \ref{th:main}, and $Q = \frac{\sqrt{\rho}M}{\sqrt{\epsilon}L}$, Algorithm \ref{alg:cap} will return $(\epsilon, \sqrt{\rho\epsilon})$-SOSP after observing $\tilde{O}(\frac{1}{\epsilon^3})$ trajectories. Thus, it improves the best-known sample complexity in \citep{wang2022stochastic, maniyar2024cubic} by a factor of $O(\epsilon^{-0.5})$. 
\end{corollary}

\section{Related Work}
\label{sec:rel_work}
Variance-reduced PG methods have been proposed in the RL literature to reduce the variance of the stochastic gradient and improve the training process \citep{papini2018stochastic, xu2019sample, zhang2021convergence}. 
Variance-reduced methods such as SARAH \citep{nguyen2017sarah}, SAGA \citep{defazio2014saga}, and SVRG \citep{ johnson2013accelerating} proposed in the context of stochastic optimization, are the basis of some more recent variance-reduced policy gradient methods in RL. 
These approaches use the difference $\nabla f(\theta',z)-\nabla f(\theta,z)$ for two consecutive points $\theta$ and $\theta'$ along iterations for the same randomness $z$ to reduce the variance of gradient estimates. This difference can be easily computed in tasks such as supervised learning as the randomness $z$ does not depend on the parameters to be optimized (e.g., parameters of the decision rule in supervised learning).
However, in the RL setting, the distribution over trajectories depends on the parameters of the current policy. Thus, most variance-reduced methods in RL require IS weights in order to provide unbiased estimates of $\nabla J(\theta)$ for a given point $\theta$ based on a trajectory that is generated by a policy with another parameter $\theta'$. As a result, the convergence analysis for these methods requires strong assumptions such as the boundedness of variance of IS weights. Examples of methods requiring such strong assumptions are SRVR-PG \citep{xu2020improved}, ProxHSPGA \citep{pham2020hybrid}, IS-MBPG \citep{huang2020momentum}, and PAGE-PG \citep{gargiani2022page} which all of them converge to an $\epsilon$-FOSP with the sample complexity of $O(\epsilon^{-3})$. HAPG is the first work achieving the same rate by using second-order information (in the form of HVP) to bypass IS weights. 
However, in their analysis, it is required to use a fixed step size of $\epsilon$ which slows the training process in practice. Moreover, the number of computed HVPs per iteration is in the order of $O(1/\epsilon)$.
Very recently, two methods using mirror-descent algorithm based on Bregman divergence, called VR-MPO \citep{yang2022policy} and VR-BGPO \citep{huang2021bregman} have been proposed. These methods achieve $\epsilon$-FOSP if the mirror map in Bregman divergence is the $l_2$-norm.


In order to converge to SOSP, in the context of optimization, \citet{nesterov2006cubic} analyzed the cubic-regularized Newton (CRN) method in the deterministic case. In this case, a sub-problem  formulated based on a second-order Taylor expansion of the objective function with a cubic penalty term is solved at each iteration. CRN uses the full Hessian matrix in the minimization of sub-problem which is not practical in large-scale applications. More recently, \citet{carmon2016gradient} proposed a gradient descent method to find $\epsilon$-global optimum point of the sub-problem. The proposed method only requires computing HVPs (without requiring the full Hessian matrix) which can be computed with the same computational complexity of obtaining stochastic gradients \citep{pearlmutter1994fast}. 
In the stochastic setting, \citet{tripuraneni2018stochastic} proposed stochastic CRN (SCRN) that uses sub-sampled gradients and Hessian vector products to solve the sub-problem at each iteration and converges to $(\epsilon, \sqrt{\rho\epsilon})$-SOSP with sample complexity of $\tilde{O}(\epsilon^{-3.5})$. 
Later, \citet{zhou2020stochastic} introduced a variance-reduced version of SCRN with sample complexity of $\tilde{O}(\epsilon^{-3})$.
{In the context of RL, \citet{yang2021sample} analyzed REINFORCE under some restrictive assumptions on the objective function and showed that it convergences to $(\epsilon,\sqrt{\rho\epsilon})$-SOSP with sample complexity of $\tilde{O}(\epsilon^{-4.5})$.  Recently, \citet{wang2022stochastic} introduced a stochastic cubic-regularized policy gradient method that achieves $(\epsilon, \sqrt{\rho\epsilon})$-SOSP with the sample complexity of $\tilde{O}(\epsilon^{-3.5})$. 
This method still requires IS weights and consequently strong assumptions about them. Concurrent to \citet{wang2022stochastic}, \citet{maniyar2024cubic} proposed ACR-PN, a second-order method which achieves $(\epsilon, \sqrt{\rho\epsilon})$-SOSP with a sample complexity of $O(\epsilon^{-3.5})$.}
Our proposal, VR-SCP algorithm, does not require IS and achieves $(\epsilon, \sqrt{\rho\epsilon})$-SOSP with sample complexity of $\tilde{O}(\epsilon^{-3})$. To the best of our knowledge, VR-SCP is the first algorithm in the literature to provide such a guarantee.

\section{Experiments}
\label{sec:exp}

In this section, we evaluate the proposed algorithm and compare it with the related work in four control tasks in MuJoCo  \citep{todorov2012mujoco} which is a physics simulator, with fast and accurate simulations in areas such as robotics, bio-mechanics, graphics, etc. We used
Garage library \citep{garage} for our implementations as it allows for maintaining and integrating several RL algorithms\footnote{The code for all experiments is available at \url{https://github.com/sadegh16/VR-SCP}.}.

In the following, we briefly explain the control tasks in the four environments we consider. 
In Walker environment, a humanoid walker tries to move forward in a two-dimensional space. It can only fall forward or backward and the goal is to walk as long as possible without falling.
In Reacher environment, there is an arm that tries to reach a specific point in a plane and the goal is to navigate the arm such that the tip of the arm gets as close as possible to that point.
In Hopper environment, there is a two-dimensional one-legged robot and the goal is to make it hop in the forward (right) direction as long as possible.
In Humanoid environment, there is a 3D bipedal robot like a human and the goal is to make it walk forward as fast as possible without falling over.

We compared our proposed algorithm with PG methods that provide theoretical guarantees: 
PAGE-PG \citep{gargiani2022page}, IS-MBPG \citep{huang2020momentum} which is based on STROM, HAPG \citep{shen2019hessian} which does not require IS weights, VR-BGPO \citep{huang2021bregman} which is a mirror descent based algorithm, {and ACR-PN \citep{maniyar2024cubic} which is the state-of-the-art of second-order methods in the RL setting.}
These algorithms have guarantees on convergence to an approximate FOSP. We also considered REINFORCE \citep{sutton2000policy} as a baseline algorithm.
There are some other approaches in the literature with theoretical guarantees: (SCR-PG \citep{wang2022stochastic},  VRMPO \citep{yang2022policy}, and STORM-PG \citep{ding2021global}) but the official implementations are not publicly available and our request to access the code from the authors remained unanswered.

For each algorithm, we use the same set of Gaussian policies parameterized with neural networks consisting of two layers of 64 neurons each. 
Baselines and environment settings (such as maximum trajectory horizon, and reward scale) are considered the same for all algorithms. We chose a maximum horizon of 500 for Walker, Hopper, and Humanoid and 50 for Reacher. More details about experiments are given in Appendix \ref{apdx:exp}. It is worth mentioning that we used the official implementation of each algorithm. For REINFORCE, we used the implementation provided by Garage library \citep{garage}. For PAGE-PG, we found some issues in the official code and we decided to carefully implement it following the description of the original paper. The implementation of VR-SCP (our algorithm) is available as supplementary material.

There are two main challenges in evaluating PG methods experimentally. First, it has been observed that PG methods are often sensitive to parameter initialization and random seeds \citep{henderson2018deep}. Thus, it is quite difficult in some cases to reproduce previous results.
Second, it is unclear how to compare algorithms in terms of performance (e.g., the average return over instances of an algorithm) and robustness (e.g., the standard deviation of return (STD return) over instances of an algorithm) simultaneously.
For instance, between an algorithm, with both a high average return and STD return and another algorithm with a lower average return but with also a lower STD return, which one is preferable?
\begin{figure}[htp!]
    \centering
    \includegraphics[width=.3\textwidth]{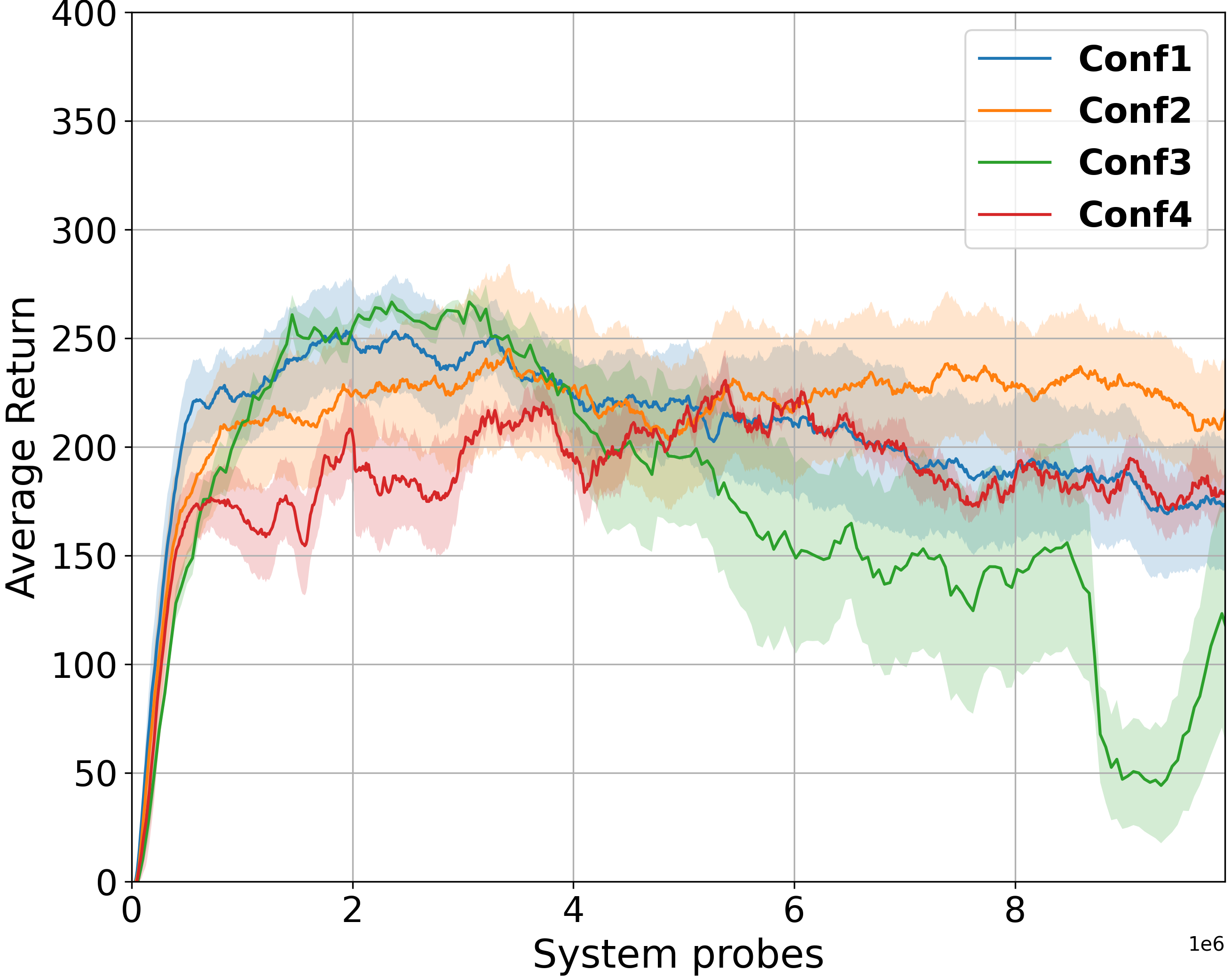}
    \caption{
    Each configuration is evaluated with five different random seeds.}
    \label{fig:hyper}
\end{figure}

In this section, we aim to find a metric to simultaneously capture the average return as well as STD return of an algorithm. Additionally, we would like to assess an algorithm's sensitivity to random seeds, which is central for reproducibility. 
To address the first question, we define the following metric. For any algorithm $A$, after observing $t$ number of state-actions pairs (called system probes), we compute the lower bound of the confidence interval of average return over $n$ runs of the algorithm and denote it by $LCI_A(n,t)$.
We define the performance-robustness (PR) metric  by averaging $LCI_A(n,t)$ over all system probes $t=1,\cdots,T$ as follows:
\begin{equation}
    \label{eq:PM}
    PR_A(n) = \dfrac{1}{T}\sum_{t=1}^{T} LCI_A(n,t),
\end{equation}
where $T$ is the maximum number of system probes.
\begin{figure*}[htp!]
    \centering
    \includegraphics[width=0.68\textwidth]{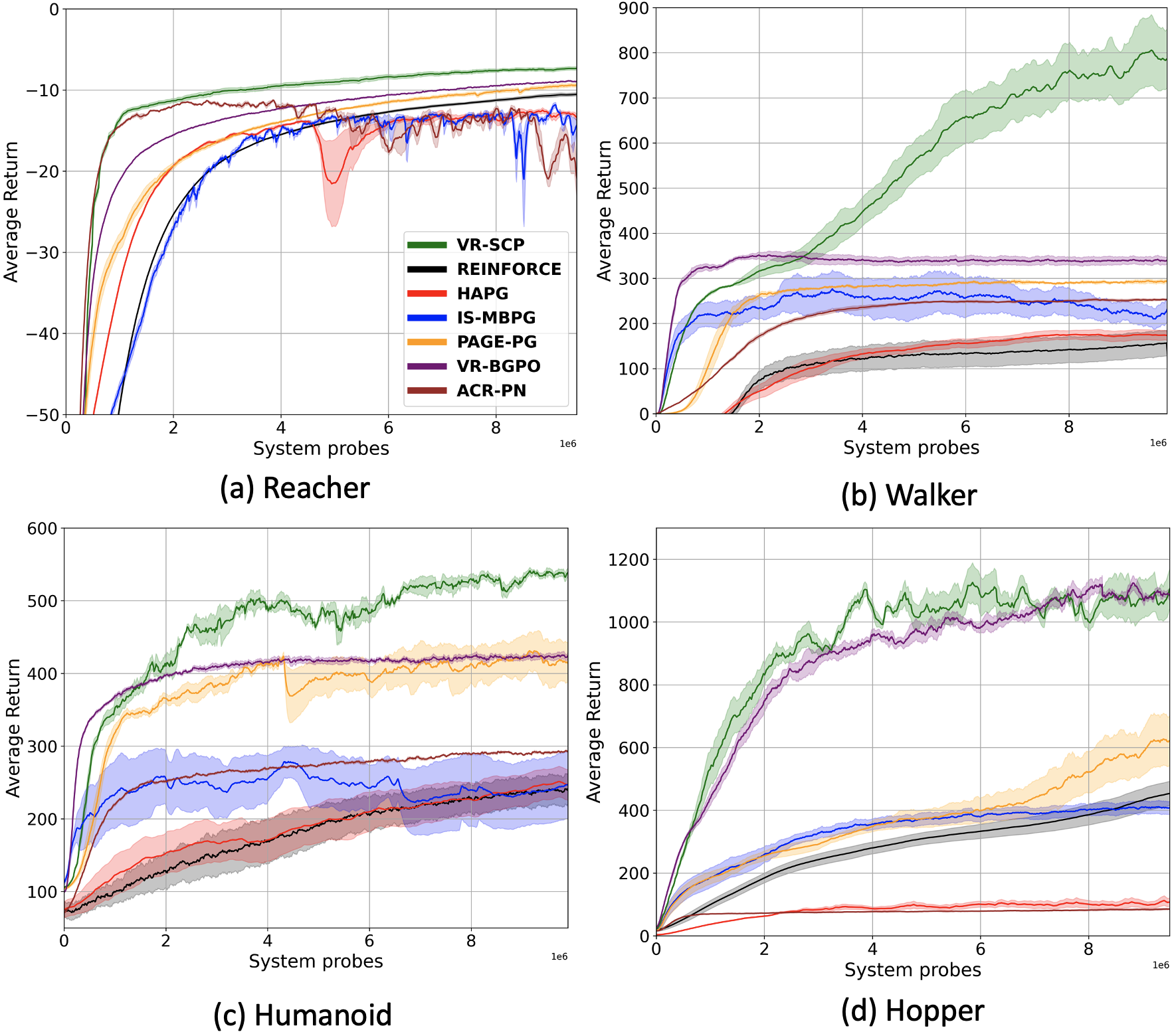}
    \caption{Comparison of VR-SCP with other variance reduction methods on four control tasks. }
    \label{fig:qualitative}
\end{figure*}
\begin{table}[t]
    \centering
    \caption{Comparison of VR-SCP with other variance-reduced methods in terms of PR. In each environment, the highest PR is in bold. } 

    \renewcommand{\arraystretch}{2}
    \resizebox{1\columnwidth}{!}{%
    \begin{tabular}{|c|c|c|c|c|} 
        \hline 
        & Reacher & Walker & Humanoid & Hopper\\\hline 
            VR-SCP (our algorithm) & \textbf{-10.88} & \textbf{486.08} & \textbf{484.19} &  \textbf{891.48}\\\hline
        HAPG & -19.51 & 104.296 & 161.12& 74.60\\\hline
        IS-MBPG & -21.76& 204.32&  201.01& 312.21\\\hline
        PAGE-PG & -17.39 & 247.58& 356.58&  338.72\\\hline
        REINFORCE & -20.10 & 79.98 & 156.21& 263.92 \\\hline
        VR-BGPO & -15.15 & 320.51&  409.67&  861.92\\\hline
        ACR-PN & -17.19 & 200.84&  260.09&  74.05\\\hline
        
    \end{tabular}
    }
    \label{table: runs}
\end{table}
Figure \ref{fig:hyper}  illustrates why we take the average of $LCI_A(n,t)$ over all system probes $t=1,\cdots,T$ in the definition of our PR metric. In this figure,  four different configurations of hyper-parameters for IS-MBPG algorithm \citep{huang2020momentum} are considered in Walker environment. PR values for these configurations are $234.8, 246.9, 193.5,$ and $199.1$, respectively. Taking the average of $LCI_A(n,t)$ over the probes prevents us from choosing the hyper-parameters that aggressively improve the returns in the beginning but can degrade drastically by the end of the horizon.  

We used grid search to tune the hyper-parameters of all the algorithms. For the algorithms except ours, the search space for each hyper-parameter was chosen based on the one from the original papers. For each configuration of the hyper-parameters, we ran each algorithm $A$, five times and computed $PR_A(5)$. We selected the configuration that maximized $PR_A(5)$ and then reported $PR_A(10)$ of each algorithm for the selected configuration based on 10 different runs in Table \ref{table: runs}. Our proposed method achieved the highest PR in all environments compared to the other algorithms. 

To address the second question, we consider the confidence interval of the performance to gauge the sensitivity of an algorithm to random seeds. 
In Figure \ref{fig:qualitative}, VR-SCP exhibits better performance and robustness compared to the other algorithms. It has relatively less variance as well as a higher average return. This could be explained by the fact that as VR-SCP converges to an $(\epsilon,\sqrt{\rho\epsilon})$-SOPS, it is less sensitive to random seeds which results in smaller confidence intervals. Additionally, as it avoids saddle points and possibly bad local optima it achieves a higher average return. This is most evident in Walker environment, where the other algorithms get stuck in a saddle point or local maxima but VR-SCP escapes them.
It is worth mentioning that, in our experiments, REINFORCE has comparable performance to some variance-reduced PG methods, while in previous work, the performance of REINFORCE was often reported much worse, which might have been due to poor tuning of its hyper-parameters. 

\section{Conclusion}
\label{sec:con}
We proposed a variance-reduced cubic regularized Newton PG method that uses second-order information in the form of HVP and converges to $(\epsilon,\sqrt{\rho\epsilon})$-SOSP with the sample complexity of $\Tilde{O}(\epsilon^{-3})$.
Such a guarantee ensures escaping saddle points and bad local optima. Our rate improves the best-known sample complexity by a factor of $O(\epsilon^{-0.5})$. Moreover, unlike most methods in the literature, we do not require IS weights in the variance reduction part. In the experiments, due to the sensitivity of the algorithms to random seeds,
we defined a new measure to account simultaneously for the performance (average return) and robustness (standard deviation of return) of an algorithm and used it in our evaluations.
Experimental results clearly showcase the advantage of VR-SCP both in terms of its performance and robustness in a variety of control tasks compared with other methods in the literature.

\newpage



\begin{acknowledgements} 
This research was supported by the Swiss National Science Foundation through the NCCR Automation program under grant agreement no. 51NF40\_180545.
\end{acknowledgements}

\bibliography{camera-ready}

\begin{thebibliography}{38}
\providecommand{\natexlab}[1]{#1}
\providecommand{\url}[1]{\texttt{#1}}
\expandafter\ifx\csname urlstyle\endcsname\relax
  \providecommand{\doi}[1]{doi: #1}\else
  \providecommand{\doi}{doi: \begingroup \urlstyle{rm}\Url}\fi

\bibitem[Baxter and Bartlett(2001)]{baxter2001infinite}
Jonathan Baxter and Peter~L Bartlett.
\newblock Infinite-horizon policy-gradient estimation.
\newblock \emph{Journal of Artificial Intelligence Research}, 15:\penalty0 319--350, 2001.

\bibitem[Carmon and Duchi(2016)]{carmon2016gradient}
Yair Carmon and John~C Duchi.
\newblock Gradient descent efficiently finds the cubic-regularized non-convex newton step.
\newblock \emph{arXiv preprint arXiv:1612.00547}, 2016.

\bibitem[Defazio et~al.(2014)Defazio, Bach, and Lacoste-Julien]{defazio2014saga}
Aaron Defazio, Francis Bach, and Simon Lacoste-Julien.
\newblock Saga: A fast incremental gradient method with support for non-strongly convex composite objectives.
\newblock \emph{Advances in neural information processing systems}, 27, 2014.

\bibitem[Deisenroth et~al.(2013)Deisenroth, Neumann, Peters, et~al.]{deisenroth2013survey}
Marc~Peter Deisenroth, Gerhard Neumann, Jan Peters, et~al.
\newblock A survey on policy search for robotics.
\newblock \emph{Foundations and trends in Robotics}, 2\penalty0 (1-2):\penalty0 388--403, 2013.

\bibitem[Ding et~al.(2021)Ding, Zhang, and Lavaei]{ding2021global}
Yuhao Ding, Junzi Zhang, and Javad Lavaei.
\newblock On the global convergence of momentum-based policy gradient.
\newblock \emph{arXiv preprint arXiv:2110.10116}, 2021.

\bibitem[garage contributors(2019)]{garage}
The garage contributors.
\newblock Garage: A toolkit for reproducible reinforcement learning research.
\newblock \url{https://github.com/rlworkgroup/garage}, 2019.

\bibitem[Gargiani et~al.(2022)Gargiani, Zanelli, Martinelli, Summers, and Lygeros]{gargiani2022page}
Matilde Gargiani, Andrea Zanelli, Andrea Martinelli, Tyler Summers, and John Lygeros.
\newblock Page-pg: A simple and loopless variance-reduced policy gradient method with probabilistic gradient estimation.
\newblock \emph{arXiv preprint arXiv:2202.00308}, 2022.

\bibitem[Henderson et~al.(2018)Henderson, Islam, Bachman, Pineau, Precup, and Meger]{henderson2018deep}
Peter Henderson, Riashat Islam, Philip Bachman, Joelle Pineau, Doina Precup, and David Meger.
\newblock Deep reinforcement learning that matters.
\newblock In \emph{Proceedings of the AAAI conference on artificial intelligence}, volume~32, 2018.

\bibitem[Huang et~al.(2020)Huang, Gao, Pei, and Huang]{huang2020momentum}
Feihu Huang, Shangqian Gao, Jian Pei, and Heng Huang.
\newblock Momentum-based policy gradient methods.
\newblock In \emph{International Conference on Machine Learning}, pages 4422--4433. PMLR, 2020.

\bibitem[Huang et~al.(2021)Huang, Gao, and Huang]{huang2021bregman}
Feihu Huang, Shangqian Gao, and Heng Huang.
\newblock Bregman gradient policy optimization.
\newblock In \emph{International Conference on Learning Representations}, 2021.

\bibitem[Johnson and Zhang(2013)]{johnson2013accelerating}
Rie Johnson and Tong Zhang.
\newblock Accelerating stochastic gradient descent using predictive variance reduction.
\newblock \emph{Advances in neural information processing systems}, 26:\penalty0 315--323, 2013.

\bibitem[Konda and Tsitsiklis(2000)]{konda2000actor}
Vijay~R Konda and John~N Tsitsiklis.
\newblock Actor-critic algorithms.
\newblock In \emph{Advances in neural information processing systems}, pages 1008--1014, 2000.

\bibitem[Maniyar et~al.(2024)Maniyar, Prashanth, Mondal, and Bhatnagar]{maniyar2024cubic}
Mizhaan~P Maniyar, LA~Prashanth, Akash Mondal, and Shalabh Bhatnagar.
\newblock A cubic-regularized policy newton algorithm for reinforcement learning.
\newblock In \emph{International Conference on Artificial Intelligence and Statistics}, pages 4708--4716. PMLR, 2024.

\bibitem[Masiha et~al.(2022)Masiha, Salehkaleybar, He, Kiyavash, and Thiran]{masiha2022stochastic}
Saeed Masiha, Saber Salehkaleybar, Niao He, Negar Kiyavash, and Patrick Thiran.
\newblock Stochastic second-order methods provably beat sgd for gradient-dominated functions.
\newblock In \emph{Advances in neural information processing systems}, 2022.

\bibitem[Nesterov and Polyak(2006)]{nesterov2006cubic}
Yurii Nesterov and Boris~T Polyak.
\newblock Cubic regularization of newton method and its global performance.
\newblock \emph{Mathematical Programming}, 108\penalty0 (1):\penalty0 177--205, 2006.

\bibitem[Nguyen et~al.(2017)Nguyen, Liu, Scheinberg, and Tak{\'a}{\v{c}}]{nguyen2017sarah}
Lam~M Nguyen, Jie Liu, Katya Scheinberg, and Martin Tak{\'a}{\v{c}}.
\newblock Sarah: A novel method for machine learning problems using stochastic recursive gradient.
\newblock In \emph{International Conference on Machine Learning}, pages 2613--2621. PMLR, 2017.

\bibitem[Papini et~al.(2018)Papini, Binaghi, Canonaco, Pirotta, and Restelli]{papini2018stochastic}
Matteo Papini, Damiano Binaghi, Giuseppe Canonaco, Matteo Pirotta, and Marcello Restelli.
\newblock Stochastic variance-reduced policy gradient.
\newblock In \emph{International conference on machine learning}, pages 4026--4035. PMLR, 2018.

\bibitem[Pearlmutter(1994)]{pearlmutter1994fast}
Barak~A Pearlmutter.
\newblock Fast exact multiplication by the hessian.
\newblock \emph{Neural computation}, 6\penalty0 (1):\penalty0 147--160, 1994.

\bibitem[Pham et~al.(2020)Pham, Nguyen, Phan, Nguyen, Dijk, and Tran-Dinh]{pham2020hybrid}
Nhan Pham, Lam Nguyen, Dzung Phan, Phuong~Ha Nguyen, Marten Dijk, and Quoc Tran-Dinh.
\newblock A hybrid stochastic policy gradient algorithm for reinforcement learning.
\newblock In \emph{International Conference on Artificial Intelligence and Statistics}, pages 374--385. PMLR, 2020.

\bibitem[Pirotta et~al.(2013)Pirotta, Restelli, and Bascetta]{pirotta2013adaptive}
Matteo Pirotta, Marcello Restelli, and Luca Bascetta.
\newblock Adaptive step-size for policy gradient methods.
\newblock \emph{Advances in Neural Information Processing Systems}, 26, 2013.

\bibitem[Schulman et~al.(2015{\natexlab{a}})Schulman, Levine, Abbeel, Jordan, and Moritz]{schulman2015trust}
John Schulman, Sergey Levine, Pieter Abbeel, Michael Jordan, and Philipp Moritz.
\newblock Trust region policy optimization.
\newblock In \emph{International conference on machine learning}, pages 1889--1897. PMLR, 2015{\natexlab{a}}.

\bibitem[Schulman et~al.(2015{\natexlab{b}})Schulman, Moritz, Levine, Jordan, and Abbeel]{schulman2015high}
John Schulman, Philipp Moritz, Sergey Levine, Michael Jordan, and Pieter Abbeel.
\newblock High-dimensional continuous control using generalized advantage estimation.
\newblock \emph{arXiv preprint arXiv:1506.02438}, 2015{\natexlab{b}}.

\bibitem[Schulman et~al.(2017)Schulman, Wolski, Dhariwal, Radford, and Klimov]{schulman2017proximal}
John Schulman, Filip Wolski, Prafulla Dhariwal, Alec Radford, and Oleg Klimov.
\newblock Proximal policy optimization algorithms.
\newblock \emph{arXiv preprint arXiv:1707.06347}, 2017.

\bibitem[Shalev-Shwartz et~al.(2016)Shalev-Shwartz, Shammah, and Shashua]{shalev2016safe}
Shai Shalev-Shwartz, Shaked Shammah, and Amnon Shashua.
\newblock Safe, multi-agent, reinforcement learning for autonomous driving.
\newblock \emph{arXiv preprint arXiv:1610.03295}, 2016.

\bibitem[Shen et~al.(2019)Shen, Ribeiro, Hassani, Qian, and Mi]{shen2019hessian}
Zebang Shen, Alejandro Ribeiro, Hamed Hassani, Hui Qian, and Chao Mi.
\newblock Hessian aided policy gradient.
\newblock In \emph{International conference on machine learning}, pages 5729--5738. PMLR, 2019.

\bibitem[Silver et~al.(2017)Silver, Schrittwieser, Simonyan, Antonoglou, Huang, Guez, Hubert, Baker, Lai, Bolton, et~al.]{silver2017mastering}
David Silver, Julian Schrittwieser, Karen Simonyan, Ioannis Antonoglou, Aja Huang, Arthur Guez, Thomas Hubert, Lucas Baker, Matthew Lai, Adrian Bolton, et~al.
\newblock Mastering the game of go without human knowledge.
\newblock \emph{nature}, 550\penalty0 (7676):\penalty0 354--359, 2017.

\bibitem[Sutton et~al.(2000)Sutton, McAllester, Singh, and Mansour]{sutton2000policy}
Richard~S Sutton, David~A McAllester, Satinder~P Singh, and Yishay Mansour.
\newblock Policy gradient methods for reinforcement learning with function approximation.
\newblock In \emph{Advances in neural information processing systems}, pages 1057--1063, 2000.

\bibitem[Todorov et~al.(2012)Todorov, Erez, and Tassa]{todorov2012mujoco}
Emanuel Todorov, Tom Erez, and Yuval Tassa.
\newblock Mujoco: A physics engine for model-based control.
\newblock In \emph{2012 IEEE/RSJ International Conference on Intelligent Robots and Systems}, pages 5026--5033. IEEE, 2012.

\bibitem[Tripuraneni et~al.(2018)Tripuraneni, Stern, Jin, Regier, and Jordan]{tripuraneni2018stochastic}
Nilesh Tripuraneni, Mitchell Stern, Chi Jin, Jeffrey Regier, and Michael~I Jordan.
\newblock Stochastic cubic regularization for fast nonconvex optimization.
\newblock \emph{Advances in neural information processing systems}, 31, 2018.

\bibitem[Tropp(2012)]{tropp2012user}
Joel~A Tropp.
\newblock User-friendly tail bounds for sums of random matrices.
\newblock \emph{Foundations of computational mathematics}, 12:\penalty0 389--434, 2012.

\bibitem[Wang et~al.(2022)Wang, Wang, and Zheng]{wang2022stochastic}
Pengfei Wang, Hongyu Wang, and Nenggan Zheng.
\newblock Stochastic cubic-regularized policy gradient method.
\newblock \emph{Knowledge-Based Systems}, 255:\penalty0 109687, 2022.

\bibitem[Williams(1992)]{williams1992simple}
Ronald~J Williams.
\newblock Simple statistical gradient-following algorithms for connectionist reinforcement learning.
\newblock \emph{Machine learning}, 8\penalty0 (3):\penalty0 229--256, 1992.

\bibitem[Xu et~al.(2019)Xu, Gao, and Gu]{xu2019sample}
Pan Xu, Felicia Gao, and Quanquan Gu.
\newblock Sample efficient policy gradient methods with recursive variance reduction.
\newblock \emph{arXiv preprint arXiv:1909.08610}, 2019.

\bibitem[Xu et~al.(2020)Xu, Gao, and Gu]{xu2020improved}
Pan Xu, Felicia Gao, and Quanquan Gu.
\newblock An improved convergence analysis of stochastic variance-reduced policy gradient.
\newblock In \emph{Uncertainty in Artificial Intelligence}, pages 541--551. PMLR, 2020.

\bibitem[Yang et~al.(2021)Yang, Zheng, and Pan]{yang2021sample}
Long Yang, Qian Zheng, and Gang Pan.
\newblock Sample complexity of policy gradient finding second-order stationary points.
\newblock In \emph{Proceedings of the AAAI Conference on Artificial Intelligence}, volume~35, pages 10630--10638, 2021.

\bibitem[Yang et~al.(2022)Yang, Zhang, Zheng, Zheng, Li, Huang, and Pan]{yang2022policy}
Long Yang, Yu~Zhang, Gang Zheng, Qian Zheng, Pengfei Li, Jianhang Huang, and Gang Pan.
\newblock Policy optimization with stochastic mirror descent.
\newblock In \emph{Proceedings of the AAAI Conference on Artificial Intelligence}, volume~36, pages 8823--8831, 2022.

\bibitem[Zhang et~al.(2021)Zhang, Ni, Yu, Szepesvari, and Wang]{zhang2021convergence}
Junyu Zhang, Chengzhuo Ni, Zheng Yu, Csaba Szepesvari, and Mengdi Wang.
\newblock On the convergence and sample efficiency of variance-reduced policy gradient method.
\newblock \emph{arXiv preprint arXiv:2102.08607}, 2021.

\bibitem[Zhou and Gu(2020)]{zhou2020stochastic}
Dongruo Zhou and Quanquan Gu.
\newblock Stochastic recursive variance-reduced cubic regularization methods.
\newblock In \emph{International Conference on Artificial Intelligence and Statistics}, pages 3980--3990. PMLR, 2020.

\end{thebibliography}

\newpage

\onecolumn

\title{Efficiently Escaping Saddle Points for Policy Optimization\\(Supplementary Material)}
\maketitle

\appendix
\section{Auxiliary Lemmas}
\begin{lemma}
\label{lem:azuma}
(Vector Azuma-Hoeffding inequality)
Let {${a_k}$} be a vector martingale difference, that is, $\mathbf{E} [a_k|\sigma(a_1, \cdots, a_{k-1})]=0$ where $\sigma(a_1, \cdots, a_{k-1})$ is $\sigma$-algebra of $a_1,\cdots, a_{k-1}$ and $\|a_k\| \leq A_k$. With probability at least $1-\delta$, 
\begin{equation}
    \left\| \sum_{k}a_k\right\| \leq 3\sqrt{\log(1/\delta)\sum_{k} A_k^2}.
\end{equation}
\end{lemma}

\begin{lemma}
\label{lem:azuma_matrix}
(\cite{tropp2012user})(Matrix Azuma inequality)
Consider a finite adapted sequence {${X_k}$} of self-adjoint matrices in dimension $d$, and a fixed sequence $\{A_k\}$ of self-adjoint matrices that $\mathbf{E} [X_k|\sigma(X_1, \cdots, X_{k-1})]=0$ where $\sigma(X_1, \cdots, X_{k-1})$ is $\sigma$-algebra of $X_1,\cdots, X_{k-1}$ and $X^2_k \preceq A^2_k$ almost surly. Then with probability at least $1-\delta$, 
\begin{equation}
    \left\| \sum_{k}X_k\right\|_2 \leq 3\sqrt{\log(d/\delta)\sum_{k} \|A_k\|_2^2}.
\end{equation}
\end{lemma}

\begin{lemma}
\label{lem:subsolver_increase}
\citep{carmon2016gradient} Consider the following function: 
\begin{equation}
    \label{eq:sub_problem}
    m(h) = \langle {v}, {h}\rangle+\frac{1}{2}\langle {Uh}, {h} \rangle-\frac{M}{6}\|{h}\|^{3},
\end{equation}
and its maximizer
\begin{equation}
\label{eq:arg_max_sub_problem}
h^* ={\bf argmax}_{{h}\in\mathbb{R}^{d}} m(h),\end{equation} 
where $v\in \mathbb{R}^d, U\in \mathbb{R}^{d\times d}$ such that $\|U\|_2 \leq L$, and $M$ is a positive constant. 

If either of the following conditions holds:\\
1- \( \sqrt{\epsilon/\rho} \leq \| h^* \| \)\\
2- \( \| v \| \geq \max \left( \frac{M\epsilon}{2\rho}, \sqrt{\frac{L M}{2}} \left( \frac{\epsilon}{\rho} \right)^{3/4} \right) \),\\

then, for \( \epsilon \leq \frac{16L^2\rho}{M^2} \) and $\mathcal{T(\epsilon)} \geq C_s \frac{L}{M\sqrt{\epsilon/\rho}}$ (where $C_s$ is a constant), with at least probability \( 1-\delta \), the Cubic Sub-solver (see Algorithm \ref{alg:subsolver}) returns an \( \hat{h} \) such that:
\[ m(\hat{h}) \geq \frac{M \rho^{-3/2} \epsilon^{3/2}}{24}.\]

\end{lemma}

\begin{lemma}
\label{lem:finalsolver_increase}
\citep{carmon2016gradient} For any $v\in \mathbb{R}^d$, and positive scalars $M$ and $L$, we define: 
\begin{align}
    R = \frac{L}{2M} +\sqrt{(\frac{L}{2M})+\|v\|/M}.
\end{align}
For $m(h)$ and $h^*$ defined in \eqref{eq:sub_problem} and \eqref{eq:arg_max_sub_problem}, respectively, 1) if $\|U\|_2 \leq L$, then $m(h)$ is ($L+2MR$)-smooth, 2) at each iteration of Cubic-Finalsolver (Algorithm \ref{alg:finalsubsolver}), $\|\Delta\| \leq \|h^*\|$ where $\Delta $ is the update vector defined in the Cubic-Finalsolver (see Algorithm \ref{alg:finalsubsolver}). 
\end{lemma}

\begin{lemma}
For any $v_t\in \mathbb{R}^{d} , U_t \in \mathbb{R}^{d\times d}, h \in \mathbb{R}^d, M\in \mathbb{R}$ such that $M/\rho \geq 2 $, we have:
\begin{align}
&\mu(\theta_t+h) \leq 9 \Bigg[M^3 \rho^{-3/2}\|h\|^3+M^{3/2}\rho^{-3/2}\Big\| \nabla J(\theta_t)-v_t\Big\|^{3/2} \\ \nonumber   
& + \rho^{-3/2}\Big\|\nabla^2 J(\theta_t)-U_t \Big\|^3 +
M^{3/2}\rho^{-3/2}\|\nabla m_t(h)\|^{3/2}+ 
M^{3/2}\rho^{-3/2} \Big|\|h\|-\|h_t^*\| \Big|^3\Bigg],
\end{align}
where $m_t(h)$ and $h^*_t$ are defined in \eqref{eq:body_sub_problem} and \eqref{eq:body_arge_max_sub_problem}, respectively. 
\end{lemma}
\noindent \textbf{Proof}:
Lemma \ref{lem:lipschizness}(c) shows the individual Lipschitz continuity of the Hessian estimator $\hat{\nabla}^{2} J(\theta,\tau)$, i.e., for any $\theta_1,\theta_2\in\mathbb{R}^d$,
\begin{equation}
    \|\hat{\nabla}^{2} J(\theta_1,\tau)-\hat{\nabla}^{2} J(\theta_2,\tau)\|\le \rho\|\theta_1-\theta_2\|_{2}.
    \label{eq:app_indLip_Hessian}
\end{equation}

Based on above inequality, we can imply the Lipschitz continuity of the Hessian ${\nabla}^{2} J(\theta)$ as follows:
\begin{equation}
\begin{split}
     \|{\nabla}^{2} J(\theta_1)-{\nabla}^{2} J(\theta_2)\|&=
    \|\mathbb{E}_{\tau}[\hat{\nabla}^{2} J(\theta_1,\tau)]-\mathbb{E}_{\tau}[\hat{\nabla}^{2} J(\theta_2,\tau)]\|\\
    &\leq\mathbb{E}_{\tau}[\|\hat{\nabla}^{2} J(\theta_1,\tau)-\hat{\nabla}^{2} J(\theta_2,\tau)\|]\\
    &\le \rho\|\theta_1-\theta_2\|_{2},
    \label{eq:app_Lip_Hessian}
\end{split}
\end{equation}
where the first inequality is due to Jensen's inequality (as the operator norm is convex) and the second inequality is according to \eqref{eq:app_indLip_Hessian}.






Let $\{\lambda_1,\cdots,\lambda_d\}$ be the eigenvalues of $\nabla^2 J(\theta_1) - \nabla^2 J(\theta_2)$.
By the definition of the operator norm:
$\|{\nabla}^{2} J(\theta_1)-{\nabla}^{2} J(\theta_2)\| = \max_i |\lambda_i|$.
According to \eqref{eq:app_Lip_Hessian}, we have:
$ \rho\|\theta_1-\theta_2\|_{2}  \geq |\lambda_i|$
for all eigenvalues \( \lambda_i \)'s. Therefore, we can imply that
$\rho \| \theta_1 - \theta_2\|\textbf{I}- \nabla^2 J(\theta_1) + \nabla^2 J(\theta_2)$ is positive semi-definite, i.e.,
\begin{equation}
    \label{eq:hessian_lipschizness_positive_semidifinite}
    \nabla^2 J(\theta_1) - \nabla^2 J(\theta_2)  \preceq \rho \| \theta_1 - \theta_2 \|\textbf{I},
\end{equation} 
where the notation $\preceq$ denotes the Loewner order, that is, for two real symmetric matrices $A$ and $B$, $A \preceq B$ if $B-A$ is positive semi-definite.

Now, we have:
\begin{align}
     \nabla^2 J(\theta_t+h) &\preceq  \nabla^2 J(\theta_t) +\rho\|h\|\textbf{I} \\
    & \preceq \Big\|\nabla^2 J(\theta_t)-U_t\Big\|\textbf{I}  +U_t +\rho\|h\|\textbf{I} \\
    & \preceq \Big\|\nabla^2 J(\theta_t)-U_t\Big\|\textbf{I} + M/2\|h_t^*\| \textbf{I} +\rho\|h\|\textbf{I}, 
\end{align}
where the first inequality is due to the Lipschitz continuity of the Hessian ${\nabla}^{2} J(\theta)$ and the second inequality holds as
$\nabla^2 J(\theta_t) -U_t \preceq \|\nabla^2 J(\theta_t) -U_t\| I$.
Moreover, the last inequality is due to the fact that the second-order derivative of $m_t(h)$, is negative semi-definite at the optimal point $h_t^*$, i.e., $U_t-M/2\|h^*_t\|\preceq 0$.
Having $\rho \leq M/2$, from the above inequality, we can imply that
\begin{align}
     \lambda_{max}(\nabla^2 J(\theta_t+h)) &\leq \Big\|\nabla^2 J(\theta_t)-U_t\Big\| + M/2\|h_t^*\|  +\rho\|h\| \\ 
    & \leq \Big\|\nabla^2 J(\theta_t)-U_t\Big\| + M\|h\|  +M\Big|\|h_t^*\|-\|h\|\Big|,
\end{align}
where we applied triangle inequality in the last step, i.e., $M/2 \|h_t^*\| \leq M/2\|h\|+M/2\big|\|h_t^*\|-\|h\|\big|\leq M/2\|h\|+M\big|\|h_t^*\|-\|h\|\big|$.
Accordingly, we have: 
\begin{align}
    & \rho^{-3/2}\lambda_{max}^3(\nabla^2 J(\theta_t+h)) \leq 9\rho^{-3/2}\Bigg[ \Big\|\nabla^2 J(\theta_t)-U_t\Big\|^3 + M^3\|h\|^3  +M^3\Big|\|h_t^*\|-\|h\|\Big|^3\Bigg],
    \label{apdx:lambdamax_bound}
\end{align}
where we used the inequality $(a+b+c)^3\leq 9(a^3+b^3+c^3)$ for any $a,b,c\geq 0$. 

Now, we obtain a bound on the norm of the gradient:
\begin{equation}
\begin{split}  \label{apdx:raw_norm_grad_bound}
     \Big\| \nabla J (\theta_t+h)\Big\|  &\leq \Big\|\nabla J (\theta_t+h)-\nabla J (\theta_t)-\nabla^2 J(\theta_t)h \Big\|+ \Big\|\nabla J (\theta_t) -v_t \Big\|\\ &\quad+\Big\|\nabla^2 J(\theta_t)h  -U_th\Big\| +\Big\|v_t +U_th-M/2\|h\|h\Big\| +M/2\|h\|^2\\
      &\overset{(a)}{=}\Big\|\nabla J (\theta_t+h)-\nabla J (\theta_t)-\nabla^2 J(\theta_t)h \Big\|+ \Big\|\nabla J (\theta_t) -v_t \Big\|\\
&\quad+\Big\|\nabla^2 J(\theta_t)h  -U_th\Big\| +\Big\|\nabla m_t(h)\Big\| +M/2\|h\|^2\\
      &\overset{(b)}{\leq}\frac{M}{4} \|h\|^2+ \Big\|\nabla J (\theta_t) -v_t \Big\|+\Big\|\nabla^2 J(\theta_t)h  -U_th\Big\| \\ 
  &\quad+\Big\|\nabla m_t(h)\Big\| +M/2\|h\|^2  \\
  &\overset{(c)}{\leq} \Big\|\nabla J (\theta_t) -v_t\Big\| +1/M\Big\|\nabla^2 J(\theta_t)  -U_t\Big\|^2 +\|\nabla m_t(h)\|+M\|h\|^2,
\end{split}
\end{equation}
(a) Due to the definition of $m_t(h)$, we have: $\Big\|v_t +U_th-M/2\|h\|h\Big \|= \|\nabla m_t(h)\|$. \\
(b) According to Hessian Lipschitzness, we have: $\Big\|\nabla J (\theta_t+h)-\nabla J (\theta_t)-\nabla^2 J(\theta_t)h \Big\| \leq 
  \rho/2\|h\|^2 \leq M/4\|h\|^2$ where we assumed that $M/\rho\geq 2$.\\
(c) We can bound the term $\|\nabla^2 J(\theta_t)h  -U_th\|$ using Cauchy-Schwarz and Young's inequalities as follows:
\begin{align}
 & \Big\|(\nabla^2 J(\theta_t)  -U_t)h\Big\| \leq 
 \Big\|\nabla^2 J(\theta_t)  -U_t\Big\|\|h\| \leq
 1/M \Big\|\nabla^2 J(\theta_t)  -U_t\Big\|^2+ M/4\|h\|^2.
\end{align}

Accordingly, we have: 
\begin{align}
    \label{apdx:norm_grad_bound}
     \| \nabla J (\theta_t+h)\|^{3/2}  \leq 2\Bigg[ \Big\|\nabla J (\theta_t) -v_t\Big\|^{3/2} +M^{-3/2}\Big\|\nabla^2 J(\theta_t)  -U_t\Big\|^3 +\|\nabla m_t(h)\|^{3/2}+M^{3/2}\|h\|^3 \Bigg],
\end{align}
where we used the inequality $(a+b+c+d)^{3/2}\leq 2(a^{3/2}+b^{3/2}+c^{3/2}+d^{3/2})$ for any $a,b,c,d\geq 0$.   

Using \eqref{apdx:lambdamax_bound} and \eqref{apdx:norm_grad_bound} and according to Definition \ref{def:inj}, we obtain an upper bound on $\mu(\theta_t+h)$:
\begin{align}
&\mu(\theta_t+h) \leq 
 9\Bigg[M^3 \rho^{-3/2}\|h\|^3+M^{3/2}\rho^{-3/2}\Big\| \nabla J(\theta_t)-v_t\Big\|^{3/2} \\ \nonumber   
& + \rho^{-3/2}\Big\|\nabla^2 J(\theta_t)-U_t \Big\|^3 +
M^{3/2}\rho^{-3/2}\|\nabla m_t(h)\|^{3/2}+ 
M^{3}\rho^{-3/2} \Big |\|h\|-\|h_t^*\| \Big|^3\Bigg],
\end{align}
which is derived by multiplying the right-hand side of \eqref{apdx:norm_grad_bound} with $ 1/8M^{3/2}\rho^{-3/2}$ (note that $ 1/8M^{3/2}\rho^{-3/2}\geq 1$ as $4\rho\leq M$) and then sum it with \eqref{apdx:lambdamax_bound}.

\begin{lemma}
\label{lem:dot_to_norm}
For any $h \in \mathbb{R}^d$, we have:
\begin{align}
    &-\frac{\rho}{8}\|h\|^3 - \frac{10}{\rho^2}\Big\|\nabla^2 J(\theta_t)-U_t \Big \|^3 \leq \Big \langle {(\nabla^2 J(\theta_{t})-U_t)h}, {h}\Big \rangle,\label{first_CY}\\
    & -\frac{\rho}{8}\|h\|^3 -\frac{6}{5\sqrt{\rho}}\Big\| \nabla J(\theta_t) -v_t\Big\|^{3/2} 
    \leq \Big\langle {\nabla J(\theta_{t})-v_t}, {h}\Big\rangle.\label{second_CY}
 \end{align}
\end{lemma}
\noindent \textbf{Proof}:
Using Cauchy-Schwarz inequality, we have:
\begin{align}
    &-\|h\|^2\Big\|\nabla^2 J(\theta_t)-U_t\Big \| \leq \Big\langle {(\nabla^2 J(\theta_{t})-U_t)h}, {h}\Big \rangle \label{Cauchy1}
    \end{align}
    On the other hand, using Young's inequality $ab\leq \frac{a^p}{p}+\frac{b^q}{q}$, where $a=(\frac{3\rho}{2^{4}})^{2/3}\|h\|^2$, $b=(\frac{2^{4}}{3\rho})^{2/3}\Big\|\nabla^2 J(\theta_t)-U_t\Big \|$, $p=3/2$ and $q=3$, we have:
    \begin{align}
    &\|h\|^2\Big\|\nabla^2 J(\theta_t)-U_t \Big\| \leq \frac{\rho}{8}\|h\|^3 + \frac{10}{\rho^2}\Big\|\nabla^2 J(\theta_t)-U_t \Big\|^3.\label{Young1}
 \end{align}
Putting \eqref{Cauchy1} and \eqref{Young1} together, we derive \eqref{first_CY}.
Similarly, we can use the following inequalities to derive \eqref{second_CY},
 \begin{align}
    &-\|h\|\Big\|\nabla J(\theta_{t})-v_t \Big \| \leq \Big\langle {\nabla J(\theta_{t})-v_t}, {h}\Big\rangle\\
    & \|h\|\Big\|\nabla J(\theta_{t})-v_t \Big\| \leq   \frac{\rho}{8}\|h\|^3 +\frac{6}{5\sqrt{\rho}}\Big\| \nabla J(\theta_t) -v_t\Big\|^{3/2},
 \end{align}
where in the second inequality, we used Young's inequality $ab\leq \frac{a^p}{p}+\frac{b^q}{q}$ where $a=(\frac{3\rho}{2^{3}})^{1/3}\|h\|$, $b=(\frac{2^{3}}{3\rho})^{1/3}\Big\|\nabla^2 J(\theta_t)-U_t\Big \|$, $p=3$ and $q=3/2$.
\begin{lemma}
\label{lem:finalsolver_max_itr}
\citep{zhang2021convergence} For $\epsilon \leq 4 L^2\rho/M$, Cubic-Finalsolver (Algorithm \ref{alg:finalsubsolver}) will terminate within  $C_F L/\sqrt{\rho\epsilon}$ iterations, where $C_F$ is a constant. 
\end{lemma}

\section{Convergence Analysis}
Consider $\xi= 4T\delta$. Let us define $\mathcal{F}_{t}$ as the history up to the point $\theta_t$ (i.e., $\sigma$-algebra of $\theta_0$ to $\theta_t$, $\mathcal{F}_{t}=\sigma(\theta_0,\theta_1, \cdots, \theta_t$)). We show the gradient estimate and sub-sampled Hessian of the objective function with $v_t$ and $U_t$, respectively.

\subsection{Proof of Lemma \ref{lem:grad_bound}}
\label{sec:lemma err est for grad}
\begin{lemma}
\label{apdx:proof_grad_bound}
Under Assumptions \ref{assum:1}, \ref{assum:2}, conditioned on $\mathcal{F}_{\OurFloor{t/Q}.Q}$, with probability at least $1-2\delta(t-\OurFloor{t/Q}.Q )$ for all $i$ where $ \OurFloor{t/Q}.Q \leq i \leq t $, we have:
    \begin{align}
       \Big \| v_i-\nabla J(\theta_i) \Big\|^2_{2}\le \frac{\epsilon^2}{30},
    \end{align}
    where for any two consecutive points $\theta_{k-1}$ and $\theta_k$ in Algorithm \ref{alg:cap}, we set 
    $S_k = C_{2}Q\frac{\|\theta_{k}-\theta_{k-1}\|_{2}^2}{\epsilon^2}$,
where $C_2= 38880L^2\log^2(1/\delta)+4^{5/4}\sqrt{1080\log(1/\delta) }\rho^2L^3_1M^{-7/4}$.


\end{lemma}

\noindent \textbf{Proof}: We have $v_t-\nabla J(\theta_t) = \sum_{k =\OurFloor{t/Q}.Q }^{t}{u_k}$
where:

\begin{equation}
    u_k=\begin{cases}
        \dfrac{1}{S_k}\sum_{s=1}^{S_k} \hat{\nabla}^2 J(\theta_{s,k},\tau_s)(\theta_k-\theta_{k-1}) - \nabla J(\theta_k) + \nabla J(\theta_{k-1}) & t \geq k > \OurFloor{t/Q}.Q\\
        \hat{\nabla} J(\theta_k, \mathcal{B}_{check})- \nabla J(\theta_k) & k = \OurFloor{t/Q}.Q,
    \end{cases}
\end{equation}


where $\theta_{s,k}=(1-\dfrac{s}{S_k})\theta_{k}+\dfrac{s}{S_k}\theta_{k-1}$.
We now find a bound on the norm of $u_k$ in both cases above.
For $k > \OurFloor{t/Q}.Q$, we rewrite $u_k$ as follows:
\begin{align}
    &u_k =\underbrace{\dfrac{1}{S_k}\sum_{s=1}^{S_k} \hat{\nabla}^2 J(\theta_{s,k},\tau_s)(\theta_k-\theta_{k-1})
    -\nabla^2 J(\theta_{s,k})(\theta_k-\theta_{k-1})}_{u_k^a}\nonumber\\
    &\qquad+\underbrace{\dfrac{1}{S_k}\sum_{s=1}^{S_k} \nabla^2 J(\theta_{s,k})(\theta_k-\theta_{k-1}) - \nabla J(\theta_k)+ \nabla J(\theta_{k-1})}_{u_k^b} .\nonumber
\end{align}
To get the upper bound on $\|u_k\|$,  we  find an upper bound on the $\|u_k^a\|$ and $\|u_k^b\|$.
For $\|u_k^a\|$, we  define  $a_s$ as follows:
    \begin{align*}
    &a_s= \hat{\nabla}^2 J(\theta_{s,k},\tau_s)(\theta_k-\theta_{k-1})
    - \nabla^2 J(\theta_{s,k})(\theta_k-\theta_{k-1}).
    \end{align*}
Due to individual gradient Lipschitzness (see Lemma \ref{lem:lipschizness}), we have: $\|\hat{\nabla}^2 J(\theta_{s,k},\tau_s)\|\leq L$. Moreover, using Jensen's inequality, $\|\nabla^2 J(\theta_{s,k})\|=\|\mathbb{E}[\hat{\nabla}^2 J(\theta_{s,k},\tau_s)]\|\leq \mathbb{E}[\|\hat{\nabla}^2 J(\theta_{s,k},\tau_s)\|]\leq L$. Therefore,
    \begin{align*}
    &\|a_s\|_{2} =\left\|\hat{\nabla}^2 J(\theta_{s,k},\tau_s)(\theta_k-\theta_{k-1})
    - \nabla^2 J(\theta_{s,k})(\theta_k-\theta_{k-1})\right\|_{2}
     \leq 2L\|\theta_{k}-\theta_{k-1}\|_{2}.
    \end{align*}

Now, using vector Azuma-Hoeffding inequality from Lemma \ref{lem:azuma}, with at least probability $1-\delta$, we have:
    \begin{align}
    &\|u_k^a\|_{2}=\left\|\dfrac{1}{S_k}\sum_{s=0}^{S_k} a_s \right\|_{2} \nonumber\\
    &\le \dfrac{3}{S_k}\sqrt{\log(1/\delta)S_k  (2L)^2\Big\|\theta_{k}-\theta_{k-1}\Big\|_{2}^2}\nonumber\\
    &\le 6L\sqrt{\dfrac{\log(1/\delta)}{S_k}}\Big\|\theta_{k}-\theta_{k-1}\Big\|_{2}. \label{ua_bound}
    \end{align}
\\
To bound $\|u_k^b\|$, we consider the following two term $o_1$ and $o_2$ in $u_k^b$:
    \begin{align}
    &\|{u_k^b}\|_{2} = \Bigg\|\underbrace{\dfrac{1}{S_k}\sum_{s=1}^{S_k} \nabla^2 J(\theta_{s,k})(\theta_k-\theta_{k-1})}_{o_1}
   \underbrace{ - \nabla J(\theta_k)+ \nabla J(\theta_{k-1})}_{o_2} \Bigg\|.
    \end{align}
    We know that $\theta_{s,k}-\theta_{s-1,k}= \Bigg [ (1-\dfrac{s}{S_k})\theta_{k}+\dfrac{s}{S_k}\theta_{k-1}\Bigg ] - \Bigg [(1-\dfrac{s-1}{S_k})\theta_{k}+\dfrac{s-1}{S_k}\theta_{k-1}\Bigg ]= \dfrac{1}{S_k}(\theta_{k-1}-\theta_{k})$. Thus, we can rewrite $o_1$ as $\sum_{s=1}^{S_k} \nabla^2 J(\theta_{s,k})(\theta_{s,k}-\theta_{s-1,k})$  and $o_2$ as $\sum_{s=0}^{S_k}
    - \nabla J(\theta_{s,k})+ \nabla J(\theta_{s-1,k})$ 
    using telescoping sum. Therefore:
    \begin{align}
     \|{u_k^b}\|_{2} &\le \sum_{s=1}^{S_k} \bigg\|-\nabla^2 J(\theta_{s,k})(\theta_{s,k}-\theta_{s-1,k})
    + \nabla J(\theta_{s,k})- \nabla J(\theta_{s-1,k})\bigg\|_{2}\nonumber\\
    &  \le {S_k} \rho \dfrac{\|\theta_{k}-\theta_{k-1}\|_{2}^2}{S_k^2}=\rho \dfrac{\|\theta_{k}-\theta_{k-1}\|_{2}^2}{S_k}, \label{ub_bound}
    \end{align}
where the second inequality is due to Hessian Lipschitzness of $J(\theta)$\footnote{Hessian Lipschitz continuity of $J$ implies that for a constant \( \rho \) and for all \( \theta_1, \theta_2  \in \mathbb{R}^{d} \): $\| \nabla J(\theta_1) - \nabla J(\theta_2) -\nabla^2 J(\theta_2)(\theta_1-\theta_2) \| \leq \rho \| \theta_1 - \theta_2 \|^2
$.}.  \\

Summing \eqref{ua_bound} and \eqref{ub_bound}, we have:
    \begin{equation}
    \|u_k\|_{2} \leq 6L\sqrt{\dfrac{\log(1/\delta)}{S_k}}\Big\|\theta_{k}-\theta_{k-1}\Big\|_{2} +
    \rho \dfrac{\Big\|\theta_{k}-\theta_{k-1}\Big\|_{2}^2}{S_k}.
    \end{equation}
Using inequality $\|A+B\|_{2}^2 \leq  2\|A\|_{2}^2 + 2\|B\|_{2}^2 $,
    \begin{equation}
    \|u_k\|_{2}^2 \leq 72L^2\dfrac{\log(1/\delta)}{S_k}\Big\|\theta_{k}-\theta_{k-1}\Big\|_{2}^2 +
    2\rho^2 \dfrac{\Big\|\theta_{k}-\theta_{k-1}\Big\|_{2}^4}{S^2_k}.
    \end{equation}

From the above inequality, we can imply that the  condition $\|u_k\|_{2}^2 \leq \dfrac{\epsilon^2}{540Q\log(1/\delta) }$ holds if $S_k$
is greater than or equal to 
\begin{equation}
  S_k\geq C_{s_1}Q\frac{\Big\|\theta_{k}-\theta_{k-1}\Big\|_{2}^2}{\epsilon^2}+C_{s_2}\sqrt{Q}\frac{\Big\|\theta_{k}-\theta_{k-1}\Big\|_{2}^2}\epsilon,  
\end{equation}

where $C_{s_1}= 38880L^2\log^2(1/\delta)$ and $ C_{s_2}= \sqrt{1080\log(1/\delta) \rho^2} $.
The above conditions has already been satisfied as we set $S_k$ to $C_{2}Q\frac{\|\theta_{k}-\theta_{k-1}\|_{2}^2}{\epsilon^2}$ in the statement of lemma where we assume that $\epsilon \leq 4 L^2\rho/M$, and $Q = \frac{\sqrt{\rho}M}{\sqrt{\epsilon}L}$.

For $k = \OurFloor{t/Q}.Q$, using Azuma-Hoeffding inequality from Lemma \ref{lem:azuma}, with probability at least $1-\delta$, we have 
\begin{align}
    &\|u_k\|_{2} =\left\|\dfrac{1}{|\mathcal{B}_{check}|}\sum_{\tau \in \mathcal{B}_{check}} \hat{\nabla} J(\theta_k, \tau)- \nabla J(\theta_k)\right\|_{2}  \\ 
    &=\dfrac{1}{|\mathcal{B}_{check}|}\left\|\sum_{\tau \in \mathcal{B}_{check}} \hat{\nabla} J(\theta_k, \tau)- \nabla J(\theta_k)\right\|_{2} \\
    &   \leq \dfrac{3}{|\mathcal{B}_{check}|}\sqrt{4\log(1/\delta) |\mathcal{B}_{check}|W^2} \\
    &\leq 3\sqrt{ 4\dfrac{\log(1/\delta)}{|\mathcal{B}_{check}|}W^2},
\end{align}
where in the first inequality, we used the inequality $\|\hat{\nabla} J(\theta_k, \tau)\|\leq W$ according to Lemma \ref{lem:lipschizness}.

Now, $ \|u_k\|^2$ can be bounded as follows:
\begin{align}
    & \|u_k\|_{2}^2 \leq 36W^2{\dfrac{\log(1/\delta)}{|\mathcal{B}_{check}|}} \\
    &  \leq \dfrac{\epsilon^2}{540\log(1/\delta) },
\end{align}
where we set $|\mathcal{B}_{check}| =\dfrac{19440W^2\log^2(1/\delta)}{\epsilon^2} $ in the second inequality.\\

Based on what we proved above, with at least probability $1-\delta(t-\OurFloor{t/Q}.Q)$, we have: $\|u_k\|_{2}^2\leq \epsilon^2/(540\log(1/\delta))$ for all $t \geq k > \OurFloor{t/Q}.Q$. Given that $\|u_k\|_{2}^2$'s are bounded by $\epsilon^2/(540\log(1/\delta))$ for all $t \geq k > \OurFloor{t/Q}.Q$, using Azuma-Hoeffding inequality, with probability at least $1-\delta$, we have:
\begin{align}
    & \| v_i-\nabla J(\theta_i) \|^2_{2} = \Big\|\sum_{k =\OurFloor{t/Q}.Q }^{i} {u_k}\Big\|_{2} ^2 \\
    & \leq 9 \log(1/\delta)\Big[(i-\OurFloor{t/Q}.Q).\dfrac{\epsilon^2}{540Q\log(1/\delta) }+ \dfrac{\epsilon^2}{540\log(1/\delta) } \Big]\\
    & \leq 9 \log(1/\delta)\Big[(t-\OurFloor{t/Q}.Q).\dfrac{\epsilon^2}{540Q\log(1/\delta) }+ \dfrac{\epsilon^2}{540\log(1/\delta) } \Big]\\
    & \leq 9\log(1/\delta).\dfrac{\epsilon^2}{270\log(1/\delta) }\\
    & \leq \frac{\epsilon^2}{30}, \label{eq:bound}
\end{align}
for any $ \OurFloor{t/Q}.Q \leq i \leq t$. 
Hence, with probability at least $1-2\delta(t-\OurFloor{t/Q}.Q)$, \eqref{eq:bound} holds for all $i$, where $ \OurFloor{t/Q}.Q \leq i \leq t $ .
\subsection{Proof of Lemma \ref{lem:hess_bound}}
\begin{lemma}
For $|\mathcal{B}_h|$ defined in the statement of Theorem \ref{th:main},
under Assumptions \ref{assum:1}, \ref{assum:2}, conditioned on $\mathcal{F}_{t}$, with probability at least $1-\delta$, we have:
    \begin{align}
        \| U_t-\nabla^2 J(\theta_t) \|^2\le \frac{\epsilon \rho}{30}.
    \end{align}
\end{lemma}

\noindent \textbf{Proof}:
Using Azuma inequality from Lemma \ref{lem:azuma_matrix}, with probability at least $1-\delta$, we have: 
\begin{align}
    &\| U_t-\nabla^2 J(\theta_t) \|= \Bigg\|\dfrac{1}{|\mathcal{B}_{h}|}\sum_{\tau \in \mathcal{B}_{h}} \hat{\nabla}^2 J(\theta_t, \tau)- \nabla ^2J(\theta_t)\Bigg\|_{2} \\ 
    &=\dfrac{1}{|\mathcal{B}_{h}|}\Bigg\|\sum_{\tau \in \mathcal{B}_{h}} \hat{\nabla} ^2 J(\theta_t, \tau)- \nabla^2 J(\theta_t)\Bigg\|_{2} \\
    & \leq \dfrac{3}{|\mathcal{B}_{h}|}\sqrt{4\log(d/\delta) |\mathcal{B}_{h}|L^2} \\
    &\leq 3\sqrt{ 4\dfrac{\log(d/\delta)}{|\mathcal{B}_{h}|}L^2},
\end{align}
where we used the the fact $\|\hat{\nabla} ^2 J(\theta_t, \tau)\|\leq L$ and $\|\nabla^2 J(\theta_t)\|\leq L$ (as shown in the proof of Lemma \ref{apdx:proof_grad_bound}).
Thus, $ \Big\| U_t-\nabla^2 J(\theta_t) \Big\|^2$ can be bounded as follows:
\begin{align}
    & \| U_t-\nabla^2 J(\theta_t) \|_{2}^2 \leq 36L^2{\dfrac{\log(d/\delta)}{|\mathcal{B}_{h}|}} \\
    & \leq \frac{\epsilon \rho}{30},
\end{align}
where we set $|\mathcal{B}_{h}| =\dfrac{1080L^2\log(d/\delta)}{\rho\epsilon} $ in the second inequality.\\

\subsection{Proof of Theorem \ref{th:main}}
Suppose that VR-SCP terminates at iteration $T^*-1$. We claim that $T^*<T$, and accordingly Cubic-Finalsolver routine is executed. By contradiction, suppose that $T^*=T$. Then for all iteration $0 \leq t \leq T^*-1$:
\begin{align}
     J(\theta_{t+1}) &\geq J(\theta_{t}) + \langle {\nabla J(\theta_{t})}, {h_t}\rangle+\frac{1}{2}\langle {\nabla^2 J(\theta_{t})h_t}, {h_t} \rangle-\frac{\rho}{6}\|{h_t}\|^{3}\\
    &  = J(\theta_{t}) + m_t(h_t) + \langle {\nabla J(\theta_{t})-v_t}, {h_t}\rangle+\frac{1}{2}\langle {(\nabla^2 J(\theta_{t})-U_t)h_t}, {h_t} \rangle+\frac{M-\rho}{6}\|{h_t}\|^{3}\\
    &  \geq J(\theta_{t}) + m_t(h_t) - \frac{6}{5\sqrt{\rho}}\Big\| \nabla J(\theta_t) -v_t\Big\|^{3/2} - \frac{10}{\rho^2}\Big\|\nabla^2 J(\theta_t)-U_t\Big \|^3 +\frac{\rho}{4}\|{h_t}\|^{3}, \label{eq:lower_bound_of_J}
\end{align}
where the equality is due to the definition of $m_t(h_t)$. Moreover,  we use Lemma \ref{lem:dot_to_norm} in the last inequality and consider $M=4\rho$.

Since we assumed that $T^*=T$, for all $0 \leq t \leq T^*-1$, we have:
\begin{align}
    \label{eq:m_t_holds}
    m_t(h_t) \geq  \rho^{-1/2}\epsilon^{3/2}/6.
\end{align}


By using Lemmas \ref{lem:grad_bound} and \ref{lem:hess_bound}, for all $0 \leq t \leq T^*-1$, with probability at least $1-3T\delta$:
\begin{align}
    &\Big\| \nabla J(\theta_t) -v_t\Big\|^{3/2} \leq \epsilon^{3/2}/20,\label{eq:grad_norm_theorem}\\
    &\Big\|\nabla^2 J(\theta_t)-U_t \Big\|^3 \leq (\rho\epsilon)^{3/2}/160.\label{eq:hess_norm_theorem}
\end{align}
Using \eqref{eq:m_t_holds}, \eqref{eq:grad_norm_theorem} and \eqref{eq:hess_norm_theorem} in \eqref{eq:lower_bound_of_J},
\begin{align}
    & J(\theta_{t+1}) - J(\theta_{t}) \geq \rho^{-1/2}\epsilon^{3/2}/6+ \frac{\rho}{4}\|{h_t}\|^{3}-\rho^{-1/2}\epsilon^{3/2}/8.
\end{align}
Telescoping the last inequality from $t=0$ to $t=T^*-1$, we have:
\begin{align}
 \Delta_J\geq J(\theta_T)-J(\theta_0) &\geq \sum_{t=0}^{T^*-1} \rho^{-1/2}\epsilon^{3/2}/24+ \frac{\rho}{4}\|{h_t}\|^{3}\label{apdx:max_deltaJ}\\
    &  \geq  T \rho^{-1/2}\epsilon^{3/2}/24+ \sum_{t=0}^{T^*-1}\frac{\rho}{4}\|{h_t}\|^{3}.
\end{align}
The last inequality contradicts with the assumption $T \geq 25\Delta_J \rho^{1/2}\epsilon^{-3/2}$. Thus, with probability at least $1-3T\delta$, Cubic-Finalsolver will be executed before ending the ``for loop'' of the Algorithm \ref{alg:cap}.

When Cubic-Finalsolver 
 is executed, according to the Lemma \ref{lem:subsolver_increase}, with probability at least $1-\delta$, none of the conditions in the statement of lemma hold. Thus we have $\sqrt{\epsilon/\rho} > \|h_t^*\|$ and $\|v_t\| < \max(M\epsilon/(2\rho), \sqrt{L M/2} (\frac{\epsilon}{\rho})^{3/4})$ with probability at least $1-T\delta$. We use this observation to bound $\mu(\theta_{\Tilde{T}}+h_{\Tilde{T}})$ where $\Tilde{T}=T^*-1$ is the last iteration. $\mu(\theta_{\Tilde{T}}+h_{\Tilde{T}})$ can be bounded as follows:
\begin{align}
    \mu(\theta_{\Tilde{T}}+h_{\Tilde{T}}) & \nonumber\leq 9\Bigg[M^3 \rho^{-3/2}\|h_{\Tilde{T}}\|^3+M^{3/2}\rho^{-3/2}\Big\| \nabla J(\theta_{\Tilde{T}})-v_{\Tilde{T}}\Big\|^{3/2} \\ \nonumber   
    & + \rho^{-3/2}\Big\|\nabla^2 J(\theta_{\Tilde{T}})-U_{\Tilde{T}} \Big\|^3 +
    M^{3/2}\rho^{-3/2}\|\nabla m_t(h_{\Tilde{T}})\|^{3/2}+ 
    M^{3}\rho^{-3/2} \Big|\|h_{\Tilde{T}}\|-\|h_{\Tilde{T}}^*\| \Big|^3 \Bigg]\\
    &  \leq 1300\epsilon^{3/2},
\end{align}
where we know $\|h_{\Tilde{T}}\| \leq \|h_{\Tilde{T}}^*\|$ using Lemma \ref{lem:finalsolver_increase}. Moreover,
 the output of Cubic-Finalsolver satisfies $\|\nabla m_t(h_{\Tilde{T}})\| \leq \epsilon$. Furthermore, we used Lemma \ref{lem:grad_bound} and \ref{lem:hess_bound} to bound $M^{3/2}\rho^{-3/2}\| \nabla J(\theta_{\Tilde{T}})-v_{\Tilde{T}}\|^{3/2}$ and $\rho^{-3/2}\|\nabla^2 J(\theta_{\Tilde{T}})-U_{\Tilde{T}} \|^3$, respectively.
\subsection{Proof of Corollary \ref{cor:main}}
We compute the number of the stochastic gradient and Hessian evaluations. First, we find a bound on the following term:
\begin{align}
    \sum_{t=0}^{\Tilde{T}} \|h_{t}\|^2 \leq (\Tilde{T}+1)^{1/3}(\sum_{t=0}^{\Tilde{T}} \|h_{t}\|^3)^{2/3} \leq  (25\Delta_J \rho^{1/2}\epsilon^{-3/2})^{1/3}+ (4\Delta_J/\rho)^{2/3}\leq  \frac{\Delta_J}{8\rho^{1/2}\epsilon^{1/2}},\label{apdx:bound_on_h_t}
\end{align}
where the first inequality is due to Hölder's inequality and the second one is due to the $\Tilde{T}=T^*-1 \leq 25\Delta_J \rho^{1/2}\epsilon^{-3/2}$ and $\Delta_J \ge  \sum_{t=0}^{T^*-1}\frac{\rho}{4}\|{h_t}\|^{3}$ according to \eqref{apdx:max_deltaJ}.
Now we compute the whole stochastic gradient evaluations over $\Tilde{T}=T^*-1$:
\begin{align}
    \sum_{\text{mod}(t,Q)=0}^{\Tilde{T}} |\mathcal{B}_{check}|+ \sum_{\text{mod}(t,Q)\neq 0}^{\Tilde{T}} S_t & \overset{(i)} \leq 
    \frac{C_1\Tilde{T}}{Q\epsilon^{2}}+ \sum_{\text{mod}(t,Q)\neq 0}^{\Tilde{T}}\frac{C_2Q\|h_{t}\|^2}{ \epsilon^{2} }\\
    & \leq \frac{C_1\Tilde{T}}{Q\epsilon^{2}}+ \frac{C_2Q}{ \epsilon^{2} } \sum_{t=0}^{\Tilde{T}} \|h_{t}\|^2\\
    & \leq \frac{C_1\Tilde{T}}{Q\epsilon^{2}}+ \frac{C_2Q}{ \epsilon^{2} } \sum_{t=0}^{\Tilde{T}} \|h_{t}\|^2\\
    & \overset{(ii)} \leq \frac{\sqrt{\epsilon}L}{\sqrt{\rho}M}\frac{C_1(25\Delta_J \rho^{1/2}\epsilon^{-3/2})}{\epsilon^{2}} + 
    \frac{C_2\sqrt{\rho}M}{ \epsilon^{2} \sqrt{\epsilon}L} \frac{\Delta_J}{8\rho^{1/2}\epsilon^{1/2}}\\
    & = \Tilde{O}(\epsilon^{-3}).
\end{align}
$(i)$ For the first term ($|\mathcal{B}_{check}|$), we consider $C_1=19440W^2\log^2(1/\delta)$. For the second term, we use $S_k$ defined in Lemma \ref{apdx:proof_grad_bound}.\\
$(ii)$ We set $Q = \frac{\sqrt{\rho}M}{\sqrt{\epsilon}L}$ and we know that $\Tilde{T}=T^*-1 \leq 25\Delta_J \rho^{1/2}\epsilon^{-3/2}$. Therefore we use \eqref{apdx:bound_on_h_t} to bound $\sum_{t=0}^{\Tilde{T}} \|h_{t}\|^2$.

Now we compute the Hessian vector evaluations in Cubic-Subslover and Cubic-Finalsolver. For each call of Cubic-Subslover, using Lemma \ref{lem:subsolver_increase}, we need $\mathcal{T(\epsilon)}\geq C_s\frac{L}{M\sqrt{\epsilon/\rho}}$ iterations and at each iteration we have $|\mathcal{B}_{h}| =\dfrac{1080L^2\log(d/\delta)}{\rho\epsilon} $ samples.
Hence in total, we have $T\times\mathcal{T(\epsilon)} \times |\mathcal{B}_{h}| = \frac{25\Delta_J \rho^{1/2}}{\epsilon^{3/2}} \times C_s\frac{L}{M\sqrt{\epsilon/\rho}} \times \dfrac{1080L^2\log(d/\delta)}{\rho\epsilon} = C_3 \epsilon^{-3} $ HVP evaluations where $C_3= 25\Delta_J C_s\frac{1}{M} 1080L^3\log(d/\delta)$.

Using Lemma \ref{lem:finalsolver_max_itr},  for Cubic-Finalsolver, the  number of HVP evaluations is as follows:

$C_F L/\sqrt{\rho\epsilon} \times \dfrac{1080L^2\log(d/\delta)}{\rho\epsilon} = C_4 \epsilon^{-3/2}$ HVP evaluations where $C_4=C_F \dfrac{1080L^3\log(d/\delta)}{\rho^{3/2}}$.
Thus in total, we have $\Tilde{O}(\epsilon^{-3})$ stochastic gradient and HVP evaluations.

\section{Descriptions of Cubic-Subsolver and Cubic-Finalsolver}

The pseudo-code of Cubic-Subsolver is given in Algorithm \ref{alg:subsolver} which uses gradient ascent to ensure that there is a sufficient increase in the objective of the sub-problem \citep{carmon2016gradient}.
One difference compared with a simple gradient ascent is that at each call of Cubic-Subsolver, if the gradient norm is large enough, it uses the Cauchy point (line 3) which guarantees sufficient increase. In the gradient ascent part, the algorithm adds a small perturbation (line 6)  (according to uniform distribution) to the gradient estimate $v$ to escape the ``hard" cases in the sub-problem and iterates for $\mathcal{T(\epsilon)}\geq C_s\frac{L}{M\sqrt{\epsilon/\rho}}$ iterations according to Lemma \ref{lem:subsolver_increase}.

Cubic-Subsolver and Cubic-Finalsolver have only access to $v$ and HVP function $U[.]$. In Cubic-Finalsolver, in the WHILE loop, the gradient ascent update is applied till the WHILE condition fails (line 2).

\label{app:subsolver}
\begin{algorithm}[t]
\caption{Cubic-Subsolver}\label{alg:subsolver}
\textbf{Input:}  $U[.], v, M, L, \epsilon$
\begin{algorithmic}[1]
\IF {$||v|| \geq L^2/M$}
        \STATE $ R_c \gets -\frac{v^TU[v]}{M||v||^2} + \sqrt{(\frac{v^TU[v]}{M||v||^2})^2 + 2||v||/M}$
        \STATE $\Delta \gets \frac{v}{||v||} R_c$
    \ELSE 
        \STATE $\Delta \gets 0 , \sigma \gets  c'\frac{\sqrt{M \epsilon}}{L} , \eta \gets \frac{1}{20L}$
        \STATE $\Tilde{v} \gets v +\sigma\mathcal{U}$  for $\mathcal{U} \sim {Uniform(S^{d-1})}$
        \FOR{$t=0,\cdots,\mathcal{T(\epsilon)}$}
        \STATE $\Delta \gets \Delta + \eta(\Tilde{v}+U[\Delta]-\frac{M}{2}||\Delta||\Delta)$
        \ENDFOR
    \ENDIF
\STATE $\Delta_m \gets v^T\Delta + \frac{1}{2}\Delta^TU[\Delta]-\frac{M}{6}||\Delta||^3$  
\STATE \textbf{return} $\Delta, \Delta_m $
\end{algorithmic}
\end{algorithm}

\begin{algorithm}[t]
\caption{Cubic-Finalsolver}\label{alg:finalsubsolver}
\textbf{Input:}  $U[.], v, M, L, \epsilon$
\begin{algorithmic}[1]
\STATE $\Delta \gets 0 , g_m \gets v , \eta \gets \frac{1}{20L}$
\WHILE {$||v_m|| \geq \epsilon/2$}
    \STATE $\Delta \gets \Delta + \eta v_m$
    \STATE $v_m \gets v+U[\Delta]-\frac{M}{2}||\Delta||\Delta$  
\ENDWHILE
\STATE \textbf{return} $\Delta$
\end{algorithmic}
\end{algorithm}

\section{Details of Experiments }
\label{apdx:exp}

We used the default implementation of linear feature baseline and Gaussian MLP baseline from Garage library. 
The employed linear feature baseline is a linear regression model that takes observations for each trajectory and extracts new features such as different powers of their lengths from the observations. These extracted features are concatenated to the observations and used to fit the parameters of the regression with the least square loss function. 

We utilized a Linux server with Intel Xeon CPU E5-2680 v3 (24 cores) operating at 2.50GHz with 377 GB DDR4 of memory and Nvidia Titan X Pascal GPU. The computation was distributed over 48 threads to ensure a relatively efficient run time.

In the following table, we provide the fine-tuned parameters for each algorithm. Batch sizes are considered the same for all algorithms. The discount factor is also set to $0.99$ for all the runs.

\begin{table*}[htp]
    \centering
    \setlength{\tabcolsep}{7pt}
    \caption{Selected hyper-parameters for different algorithms. }\renewcommand{\arraystretch}{2}
    \begin{tabular}{|c|c|c|c|c|} 
        \hline 
        & Reacher & Walker & Humanoid & Hopper\\\hline 
        Max horizon  & $50$ & $500$ & $500$ & $500$\\\hline
        Neural network sizes & $64\times64$& $64\times64$& $64\times64$& $64\times64$\\\hline
        Activation functions & Tanh& Tanh& Tanh& Tanh\\\hline
        IS-MBPG $lr$ & 0.9& 0.3& 0.75& 0.1\\\hline
        IS-MBPG $c$ & 100 & 12& 5& 50 \\\hline
        IS-MBPG $w$ & 200 & 20& 2& 100 \\\hline
        REINFORCE step-size & 0.01 & 0.01 & 0.001& 0.001 \\\hline
        VR-SCP $L$ & 200 & 50& 400& 100 \\\hline
        VR-SCP $\rho$ & 200 & 50& 50& 50 \\\hline
        VR-SCP Q & 10 & 2& 5& 2 \\\hline
        PAGE-PG $p_t$ & 0.4 & 0.4& 0.6& 0.6 \\\hline
        PAGE-PG step-size & 0.01 & 0.001& 0.0005 & 0.001 \\\hline
        HAPG step-size & 0.01 & 0.01& 0.01& 0.001 \\\hline
        HAPG $Q$ & 5 & 10& 10& 10 \\\hline
        VR-BGPO $lr$ & 0.8 & 0.75& 0.8& 0.75 \\\hline
        VR-BGPO $c$ & 25 & 25& 25& 25 \\\hline
        VR-BGPO $w$ & 1 & 1& 1& 1 \\\hline
        VR-BGPO $lam$ & 0.0005 & 0.0005& 0.0025& 0.0005 \\\hline
        ACR-PN $alpha$ & 100 & 10000& 10000& 10000 \\\hline
        
    \end{tabular}
     \label{table: grid_values}
\end{table*}

\end{document}